\documentclass[letterpaper]{article}
\usepackage[preprint,nonatbib]{nips_2018}

\usepackage{rotating}

\usepackage{times}  %Required
\usepackage{helvet}  %Required
\usepackage{courier}  %Required
\usepackage{url}  %Required
\usepackage{graphicx}  %Required
\usepackage[bottom]{footmisc}

\usepackage{booktabs}
\usepackage{hyperref}
\usepackage{subfigure}
\usepackage{icomma}
\usepackage[autostyle]{csquotes}

\usepackage{placeins}

\usepackage{wrapfig}

\usepackage{pgfplots}

\usepackage[font={footnotesize}]{caption}

\pgfmathdeclarefunction{gauss}{2}{%
  \pgfmathparse{1/(#2*sqrt(2*pi))*exp(-((x-#1)^2)/(2*#2^2))}%
}

\usepackage{algorithm}
\usepackage{algorithmic}
\usepackage{microtype}
\usepackage{eqparbox}

\usepackage{amsmath}
\usepackage{amsfonts}
\usepackage{enumitem}
\usepackage[capitalize]{cleveref}

\Crefname{section}{Sec.}{Secs.}

\usepackage[utf8]{inputenc} % allow utf-8 input
\usepackage[T1]{fontenc}    % use 8-bit T1 fonts
\usepackage{hyperref}       % hyperlinks
\usepackage{url}            % simple URL typesetting
\usepackage{booktabs}       % professional-quality tables
\usepackage{amsfonts}       % blackboard math symbols
\usepackage{nicefrac}       % compact symbols for 1/2, etc.
\usepackage{microtype}      % microtypography

\usepackage{tikz}
\usetikzlibrary{fit,positioning}
\usetikzlibrary{bayesnet}

\usetikzlibrary{chains,shapes.multipart}
\usetikzlibrary{shapes,calc}
\usetikzlibrary{automata,positioning}

\definecolor{myred}{RGB}{220,43,25}
\definecolor{mygreen}{RGB}{0,146,64}
\definecolor{myblue}{RGB}{0,143,224}

\tikzset{
myshape/.style={
  rectangle split,
  minimum height=1.5cm,
  rectangle split horizontal,
  rectangle split parts=8, 
  draw, 
  anchor=center,
  },
mytri/.style={
  draw,
  shape=isosceles triangle,
  isosceles triangle apex angle=60,
  inner xsep=0.65cm
  }
}

\title{Best-Choice Edge Grafting for Efficient Structure Learning of Markov Random Fields}

\author{Walid Chaabene\\
Department of Computer Science\\
Virginia Tech
\And
Bert Huang\\
Department of Computer Science\\
Virginia Tech}

\begin{document}

\maketitle

\begin{abstract}
Incremental methods for structure learning of pairwise Markov random fields (MRFs), such as \emph{grafting}, improve scalability by avoiding inference over the entire feature space in each optimization step. Instead, inference is performed over an incrementally grown active set of features. 
In this paper, we address key computational bottlenecks that current incremental techniques still suffer by introducing \emph{best-choice edge grafting}, an incremental, structured method that activates edges as groups of features in a streaming setting. The method uses a reservoir of edges that satisfy an activation condition, approximating the search for the optimal edge to activate. It also reorganizes the search space using search-history and structure heuristics. Experiments show a significant speedup for structure learning and a controllable trade-off between the speed and quality of learning.
\end{abstract}

\section{Introduction}

A powerful family of approaches for learning the structure of Markov random fields (MRFs) is based on minimizing $\ell_1$-regularized scores such as the negative log likelihood \cite{andrew2007scalable} of a fully connected MRF. The $\ell_1$ regularization reduces the parameters of irrelevant edges to zero.
The main challenge when using these methods is that, for large MRFs, the feature space becomes extremely large and causes an overwhelming computational cost. Active-set methods, such as grafting \cite{lee2006efficient,perkins2003grafting,zhu2010grafting}, were introduced to promote more scalability. Despite the benefits of active-set learning, grafting retains significant computational costs: As a mandatory pre-learning step, grafting computes sufficient statistics of all possible variable pairs to enable greedy activation tests on the entire search space, and each iteration of grafting requires a search over the large combinatorial space of all possible edges. 
%Furthermore, grafting does not reason about the MRF graph structure, activating single parameters at a time.

This paper introduces \emph{best-choice edge grafting}, an active-set method that activates edges in a streaming fashion as groups of parameters. The method is agnostic to the underlying inference method. We derive an edge-activation test using a structured group-$\ell_1$ learning objective. Structure learning is performed in a priority order: Edges are assigned different search priorities. We use a combination of random sampling and a min-heap priority queue \cite{van1976design} to efficiently explore the prioritized search space. The method performs "online" edge activation using a control sample of candidate edges stored in a limited-memory reservoir. Search priorities are updated based on the search history and structural information derived from the partially constructed MRF structure learned so far.  These strategies allow on-demand computation of sufficient statistics for edges that are likely to be activated, which eliminates the heavy computation of grafting's pre-learning phase. Best-choice edge grafting can start---and often finish---learning well before grafting is able to begin learning. We also introduce a trade-off parameter to balance the speed of learning and the quality of learned MRFs.  Our experiments show that best-choice edge grafting scales better to large datasets than grafting. Synthetic experiments show that the proposed method performs well at recovering the true structure, while real data experiments show that we learn high-quality MRFs from large and diverse datasets.
%These attributes lead to a significant computational speedup for structure learning of pairwise MRFs. 

\section{Background and Preliminaries}

Throughout this paper, we consider the case of log-linear, pairwise Markov random fields (MRFs). Based on a given MRF structure, the probability of a set of variables $x = \{x_1,..,x_n\}$ is
$p_w(x) = \frac{1}{Z(w)} \prod_{c \in C} \phi_c(x; w)$,
where $Z(w)$ is a normalizing partition function $Z(w) = \sum_{x} \prod_{c \in C} \phi_c(x ; w)$, and $\phi$ is a clique potential function $\phi_c(x; w) = \exp\left(\sum_{k \in c} w_k^\top f_k(x)\right)$.
The set $C$ contains sets of indices representing cliques and variables,
 and $f_k(x)$ are feature functions often defined as indicator functions. In the case of pairwise Markov random fields, a clique can refer to either a node or an edge. A pairwise MRF is associated with an undirected graph $G(V, E)$, where $V$ is the set of $n$ nodes corresponding to variables, and $E$ is the set of edges corresponding to pairwise cliques. The factorization for a pairwise MRF is 
$p_w(x) = \frac{1}{Z(w)} \prod_{i \in V} \phi_i(x; w) \prod_{(i,j) \in E} \phi_{ij}(x; w)$.

\subsection{Parameter Learning Through $\ell_1$-Regularized Likelihood}

Given a set of data $X = \{x^{(m)}\}_{ m= 1 \dots N}$, the likelihood is expressed as $l(w) = \prod_{m=1}^N p_w(x^{(m)})$.
We formulate learning as a minimization of a scaled negative log likelihood $L(w)$, where
$L(w) = - \frac{1}{N} \sum_{m=1}^N \log p_w(x^{(m)}) = - \frac{1}{N} \sum_{m=1}^N \big(w^\top f(x^{(m)}) \big) + \log Z(w)$,
and $w$ and $f$ correspond to the vectors of $w_k$ and $f_k$, respectively.  Minimizing $L(w)$ is often done using gradients. The gradient of $\log Z(W)$ with respect to the $k^{th}$ feature is the model's expectation of the $k^{th}$ feature function \cite{koller2009probabilistic}: $
\frac{\partial \log Z }{\partial w_k} = E_{w} [f_k(x)] =  \sum_{x} p_w(x) f_k(x).$
The gradient of $L$ with respect to the $k^{th}$ feature is
\begin{equation}
\begin{aligned}
\begin{split}
\frac{\partial L}{\partial w_k}  = - \frac{1}{N} \sum_{m=1}^N f_k(x^{(m)}) + E_{w} [f_k(x)] = E_{w} [f_k(x)] - E_{D} [f_k(x)] := \delta_k L.
\end{split}
\end{aligned}
\label{eq:gradient}
\end{equation}
In other words, the feature-wise gradient $\delta_k L$ is the error between the data expectation $E_{D}$ and the model expectation $E_{w}$ of $f_k(x)$. The goal of learning can be seen as minimizing that error.  To avoid over-fitting and to promote sparsity of the learned weights and of the learned Markov network, classical methods add an $\ell_1$ regularization and solve $\min_w ~ L(w) + \lambda ||w||_1$.

%This is equivalent to introducing a prior to the model parameters using a parameter distribution of the form $P(w|\lambda) = \exp(-\lambda ||w||_1)$.
%This prior will force irrelevant parameters to have zero values at the solution. As $\lambda$ increases, more parameters tend to zero out. 

\subsection{Learning Challenges}

Learning with the gradient in \cref{eq:gradient} is prohibitively expensive for three reasons: (i)~The expectation requires computing sufficient statistics for every unary and pairwise feature, which is especially expensive for large datasets. (ii)~The classical $\ell_1$ formulation ignores the structure of MRFs and treats parameters independently. This treatment dismisses the importance of local consistencies and the power of Markov independence encoded in structure. Finally, (iii)~inference of the model expectations $E_{w} [f_k(x)]$ is generally \#P-complete \cite{koller2009probabilistic}. Existing techniques focus on minimizing the cost of the inference subroutine (iii). Active-set methods allow the learning optimization to compute inference over simpler MRFs. And various approximate inference methods efficiently approximate the expectations, such as loopy belief propagation and its variants \cite{ihler2005loopy,kolmogorov2006convergent,meltzer2009convergent,murphy1999loopy,pearl2014probabilistic,wainwright2008graphical} and pseudolikelihood \cite{besag1977efficiency}. 
Group sparsity methods \cite{huang2011learning} can be straightforwardly adapted to address the structural coherence issue (ii).
Our approach extends these methods of handling issues (ii) and (iii) while also addressing the critical bottleneck of sufficient statistic computation (i) and additional bottlenecks in active-set algorithms.

\subsection{Grafting}

Grafting and its variants \cite{lee2006efficient,perkins2003grafting,zhu2010grafting} are active-set learning approaches that alternate between two primary operations:
(1) They learn parameters for an active set $S$ of features (e.g., using a sub-gradient method); and (2) they expand the active set by activating one feature using a gradient test based on the Karush-Kuhn-Tucker (KKT) conditions of the $\ell_1$-regularized objective.
Grafting converges when the activation step does not find any new feature or when a predefined maximum number of features $f_{\max}$ is reached. Grafting has a startup bottleneck with an $O(n^2 N s_{\max}^2)$ computational cost to compute the sufficient statistics for each feature, for a system of $n$ variables and a maximum number of states $s_{\max}$.
Grafting also involves an exhaustive $O(n^2 s_{\max}^2)$-time search to activate each feature. These bottlenecks translate to poor scalability for large systems and datasets.

%\emph{Grafting-light} \cite{zhu2010grafting} is a variation of grafting that runs one orthant-wise gradient step each time a new feature is selected. Although this method grows the active set faster, it requires more of the exhaustive activation tests and is more likely to produce a high number of irrelevant active features, yielding slower learning in settings with many candidate features such as MRF structure learning.

%Furthermore, it has been shown by the authors that the method suffers limitations. First, the method is very likely to introduce irrelevant features which can diverge the solution to a bad local minima. In addition, the introduction of irrelevant features can introduce an important slow down of the method and might introduce a stronger risk of over-fitting. 

Although grafting is fairly suitable for learning MRFs, it does not consider structural information or different clique-memberships of features.
%One possible outcome of grafting, or grafting light, is to activate a subset of features of an edge and keep the remaining ones at zero. This defeats the purpose of using an MRFs model which is based on the hypothesis that variables structurally influence each other.  
%Furthermore, current methods require processing the entire dataset to extract sufficient statistics tables. 
Furthermore, while grafting avoids expensive inference, it still requires various quadratic-cost operations: one that scales with the typically large amount of data and one that must be repeated each iteration of the main learning loop. 

\subsection{Group Sparsity and Edge Grafting}

Using $\ell_1$ regularization promotes sparsity uniformly across all parameters. Different regularizers can instead enforce structured sparsity \cite{huang2011learning}. A similar approach leads to a natural extension of grafting for MRF structure learning.
We refer to this variation as \emph{edge grafting}, as it activates edges instead of features. Accordingly, we define the search set $F$ and the active set $S$ respectively as sets of inactive and active edges. To include structural information in the likelihood function, we use group-$\ell_1$ regularization.  This regularization prefers parameters of each node and each edge to be homogeneous and whole edges to be sparse. We define the group-$\ell_1$ negative log likelihood as follows:
\begin{equation} \label{eq:grouplasso}
    \mathbb{L}(w) = L(w) +  \sum_{g \in G} \lambda d_{g} ||w_{g}||_2 + \lambda_{2} ||w||_2^2 ~.
\end{equation}
Each group $g$ contains the weights for either a node or an edge. Consequently, $w_{g}$ refers to the sub-vector containing all weights related to features of group $g$. We define $d_g$ as the number of states per clique. As in elastic-net methods \cite{zou2005regularization}, we add the $\ell_2$-norm to avoid some shortcomings of group-$\ell_1$ regularization, such as parameter imbalance and aggressive group selection.
We derive a KKT optimality condition:
\begin{equation} \label{eq:groupoptimality}
\begin{cases} 
     \frac{||\delta_{g} L||_2}{d_{g}} + \lambda_{2} ||w_g||_2^2 = 0 &~~~\text{ if }~~~ ||w_g||_2 \neq 0\\
     \frac{||\delta_{g} L||_2}{d_{g}} \leq \lambda &~~~\text{ if }~~~ ||w_g||_2 = 0 ~, 
   \end{cases}
\end{equation}
where $\delta_{g} L$ is the sub-vector constructed using the entries of the gradient vector $L$ corresponding to group $g$. 
From this condition, we derive an edge-activation test for a given edge $e \in F$,
\begin{equation} \label{eq:C2}
C_2 : s_e  > \lambda ~,
\end{equation}
where $s_e$ is the activation score representing the error between the model $\hat{p}_w(e)$ and the data $p_D(e)$:
\begin{equation}\label{eq:edgescore}
s_e = \tfrac{1}{d_{e}} ||\delta_{e} L||_2 = \tfrac{1}{d_e} ||\hat{p}_w(e) - p_D(e)||_2 ~.
\end{equation}
%Bert: NIPS reviewer complained that we implicitly defined d_e, but I don't see a great way to fix that
By using the group-$\ell_1$ regularizer and maintaining an active set of edges, edge grafting runs analogously to grafting but activates parameters for entire edge potentials. Edge-based group-$\ell_1$ regularization encourages structural sparsity consistent with Markov notions of variable independence. 
%With these improvements, edge grafting addresses one shortcoming of using classical grafting for MRF structure learning. However, it still requires computing sufficient statistics for all possible edges and searching over all possible edges at each iteration. 

\section{Best-Choice Edge Grafting}

Edge grafting requires computing sufficient statistics for all possible edges and searching over all possible edges at each iteration. Each of these operations costs $O(n^2)$ time. We propose \emph{best-choice edge grafting}, a method that grafts edges in a streaming fashion using a variation of reservoir sampling \cite{olken1995random}. This method activates edges without considering the entire search space. 
Best-choice edge grafting computes statistics on-demand within the learning loop, and it reorganizes the search space by assigning search priorities to edges based on search history and the structure of the partially constructed MRF graph. This reorganization helps the method test edges more likely to be relevant.  

We include pseudocode for the discussed algorithms in the appendix.

\subsection{Method Overview}

\emph{Best-choice edge grafting} activates edges without exhaustively computing all sufficient statistics or performing all activation tests. Instead, the approach starts activating edges by computing a small fraction of the edge sufficient statistics. A naive strategy would be to activate the first encountered edge that satisfies $C_2$, i.e., a ``first-hit'' approach. However, this approach can introduce many spurious edges. We propose an adaptive approach inspired by \emph{best-choice problems} that uses a control sample of edges satisfying $C_2$ stored in a limited-memory reservoir $R$. By maintaining a reservoir of potential edges to activate, we increase the probability of the algorithm activating relevant edges. (See the discussion in our complexity analysis in \cref{sec:analysis}.) Moreover, by introducing a reservoir strategy, we directly generalize both the first-hit approach and exhaustive edge grafting, which can be equivalently viewed as using size-one reservoirs and unlimited reservoirs, respectively.

A key subroutine for best-choice edge grafting selects the candidate edge with the most \emph{priority}. We define a mechanism that combines random sampling with a min-heap priority queue \cite{van1976design} to allow fast priority-based selection that requires no quadratic-time operations under mild assumptions. At all times, the edges are grouped into \emph{prioritized edges} and \emph{unseen edges}. A prioritized edge is any edge whose priority is adjusted by the algorithm, and all prioritized edges are stored in the priority queue. Unseen edges are implicitly assigned a default priority score $\rho_{0}$, but they are not explicitly stored. To select the edge with maximum priority, the algorithm first examines the maximum prioritized edge from the priority queue and compares it to $\rho_{0}$. If the edge extracted from the priority queue has lower priority than $\rho_{0}$ or if the priority queue is empty, we repeatedly sample random edges until we sample one that is not already prioritized. As we show in \cref{sec:analysis}, the probability of sampling an already prioritized edge approaches zero asymptotically as the number of variables grows. 

The learning loop selects edges in order of priority score and places edges that pass the activation test into the reservoir, which retains the edges that most violate the optimality condition. Once enough edges have been seen, the algorithm selects edges to activate from the reservoir and then performs inference to update the model expectations. The priority queue and reservoir are then updated based on the newly estimated MRF structure (see \cref{sec:reservoir,sec:priorityqueue}), and the next iteration begins. 

There are no quadratic-time operations within the main loop. Though there may in the worst case be $O(n^2)$ elements in the priority queue, each insertion, removal, or update operation costs time logarithmic in the number of stored elements, which is $O\left(\log\left(n^2\right)\right) = O(2 \log(n)) = O(\log(n))$. This efficiency enables prioritization and management of the large search space.
%We detail the specific strategies of the various phases in the remainder of this section.

\subsubsection{Reservoir management}
\label{sec:reservoir}

Best-choice edge grafting strategically manages the reservoir to retain edges likely to be relevant. Let $A \in F$ be the unknown set of all edges that currently satisfy $C_2$. We construct a control sample $R \subseteq A$ from which we activate edges. Edges in the search set $F$ are initially assigned equal, default priorities $\rho_0$. These priorities will be adjusted based on informed heuristics. The method iterates through $F$ by extracting the highest priority edge, generating a prioritized stream of edges to test. If a tested edge satisfies $C_2$, it is added to $R$. Otherwise, it is ignored. 

When a maximum number of edge tests $t_{\max}$ is reached, we start activating edges. To maintain a high-quality reservoir, edge activation only starts after we fill $R$ to capacity in the first edge-activation iteration. Furthermore, if $R$ reaches its capacity before $t_{\max}$ is reached, we replace the minimum-scoring edge in $R$ with the newly tested edge whenever it has a higher score. We use a simple strategy to activate edges in the reservoir with high activation scores where we consider edges that have at least an above-average score. 
We compute the average activation score
$\mu = \frac{1}{|R|} \sum_{e \in R} s_e$, and then we define a confidence interval for choosing edges that are likely to be relevant as
$I_{\alpha} = \big[\mu + \alpha (\max_{e\in R} s_e - \mu), \max_{e\in R} s_e \big]$,
where $\alpha \in [0,1]$. The algorithm considers activating edges with activation scores no less than
$\tau_\alpha = (1-\alpha) \mu + \alpha \max_{e\in R} s_e$. After deriving $\tau_\alpha$, edges are activated in a decreasing order with respect to their scores. To avoid redundant edges and to promote scale-free structure, we only activate edges that are not adjacent. Note that when $\alpha = 1$, we only select the maximum-scoring edge. Reducing $\alpha$ increases the number of edges to be activated at a certain step but can also result in adding spurious edges. See the pseudocode in the appendix.
%In our experiments, we show the impact of different values of $\alpha$ on the quality of the learned MRFs and convergence speed. 

After each optimization step, edge gradients change. Therefore, scores of reservoir edges are updated, and edges that no longer satisfy $C_2$ are dropped from the reservoir.

\subsubsection{Search space reorganization}
\label{sec:priorityqueue}

Best-choice edge grafting performs search space reorganization by assigning and updating the search priority of edges in the priority queue. The aim of this reorganization is to increase the quality of the received stream of edges and the reservoir. We leverage search history and structural information. 

\textbf{Search history}~~
Each activation iteration, an edge with a small activation score is unlikely to satisfy $C_2$ in the future and is placed further toward the tail of the priority queue. We define an edge-violation offset $v_{e} = 1 - \frac{s_e}{\lambda}$.
When the activation tests fail or edges are dropped from $R$, the low-score edges are not immediately returned to the priority queue but instead are ``frozen'' and placed---along with their violation offsets---in a separate container $L$. When the search priority queue is emptied, we refill it by re-injecting frozen edges from $L$ with their respective violation offsets as their new priorities.

\textbf{Partial structure information}~~
As the active set grows, so does the underlying MRF graph $G$. The resulting partial structure contains rich information about dependencies between variables. We rely on the hypothesis that graphs of real networks have a scale-free structure \cite{albert2002statistical}. We promote such structure in the learned MRF graph by encouraging testing of edges incident to central nodes. We start by measuring node centrality on the partially constructed MRF graph $G$ to detect hub nodes. A degree-based node centrality $c_i$ for a node $i$ is the fraction of all possible neighbors $N_i$ it is connected to: $c_i = |N_i| / (|V|-1)$.
We then use a centrality threshold $\hat{c}$ to identify the set of hubs $H = \{i \in V \text{ such that } c_i > \hat{c} \}$.
Finally, we prioritize all edges incident to nodes in $H$ by decrementing their priorities by 1. The total cost of updating the min-heap structure is $O(|H| n \log(n))$, where $O(\log(n))$ is the cost of updating the priority of an edge. Reorganizing the priority queue pushes edges more likely to be relevant to the front of the queue. This induces a higher-quality reservoir and promotes the activation of higher-quality edges at each activation iteration.

\subsection{Complexity Analysis}
\label{sec:analysis}

The selection of a candidate edge, either from the priority queue or from random sampling, is at most an $O(\log n)$ cost under the assumption that we only observe $O(n)$ candidate edges through learning. For each edge selection, we must examine the highest priority entry in the priority queue, which costs $O(\log n)$. Then if that priority is better than the default priority, the selection task is complete. If not, we must randomly sample an unseen edge. Randomly sampling any edge costs $O(1)$ time, and checking that the edge has not yet been seen requires another $O(1)$ set-membership check. If the edge has been seen, then we need additional samples until we sample an unseen edge. Fortunately, if the number of seen edges is less than some constant factor of $n$, i.e., $\beta n$, then the probability of sampling a previously seen edge is $\beta n /{n \choose 2}$, or $2\beta / (n-1)$, which asymptotically approaches zero. 

If it takes the algorithm $r$ edge tests to fill the reservoir in one pass, then best-choice edge grafting performs $\tilde{O}(r N s_{\max}^2)$ operations to compute the sufficient statistics necessary to fill the reservoir and activate the first edges. (We use $\tilde{O}$ notation, omitting the logarithmic costs of using the priority queue.) To activate the $j^{th}$ edge, the algorithm needs to perform at most $\tilde{O}\big((r + j t_{\max}) N s_{\max}^2\big)$ operations, where $t_{\max}$ is the allowed number of edge tests between two activation steps. 

\begin{table}[tbp]\fontsize{9}{9}\selectfont
\centering
\caption{Time complexity of different methods. The second column measures how much computation is necessary at startup. Edge grafting requires the sufficient statistics of all possible edges to do any activation, while best-choice edge grafting only computes statistics as needed. The activation-step cost is the cost to decide which edge to activate next. For reference, we include the approximate cost to approximate the expectations and gradient, which is typically linear in the size of the current active set (e.g., belief propagation or pseudolikelihood).}\label{tab:complexity}
\label{table:time}
\vspace{.1em}
\begin{tabular}{cccc}
\toprule
Algorithm & Suff. stats. at $j^{th}$ edge activation& Activation step & Inference\\
\midrule
Edge grafting & $O\big(n^2 N s_{\max}^2\big)$ & 
$O\big(n^2 s_{\max}^2\big)$ &
$\sim O(j + n)$\\ 
\addlinespace
Best-choice edge grafting & $ O\big((n + j t_{\max}) N s_{\max}^2\big)$ 
& 
$O\big(t_{\max} s_{\max}^2\big)$ &
$\sim O(j + n)$\\
\bottomrule
\end{tabular}
\end{table}

\setlength\intextsep{9pt}
\begin{wrapfigure}{O}{0.5\textwidth}
\centering
    \includegraphics[height=1.4in]{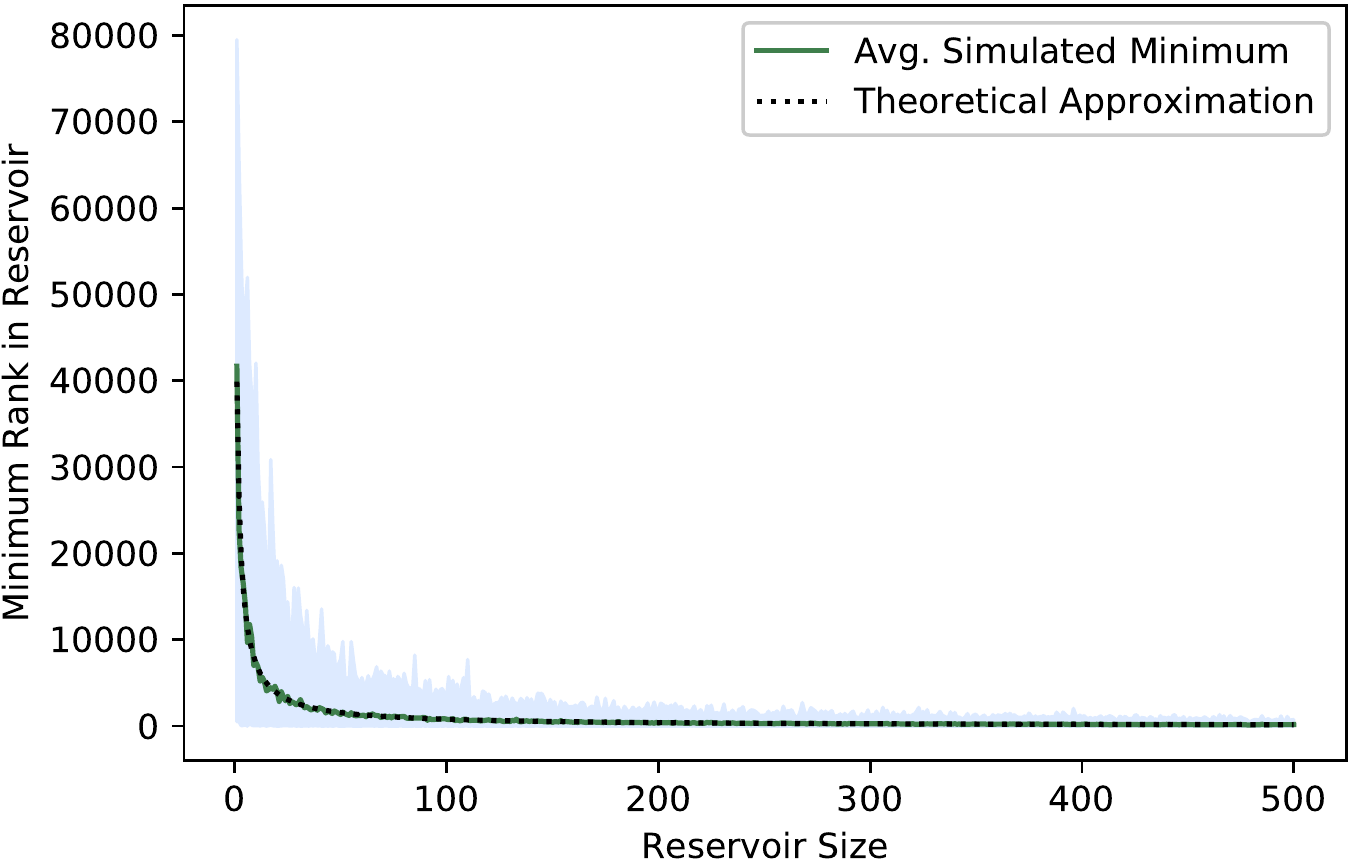}
\caption{Simulated edge ranks using the reservoir. The curves show that the average and expected ranks are nearly identical, and the shaded region indicates the full range of ranks obtained during 100 random trials.}
\label{fig:reservoir}
\end{wrapfigure}

In most cases, we can assume that it takes the algorithm $r = O(|R|)$ tests to fill the reservoir $R$. Furthermore, to construct a relevant reservoir, we set $|R| = O(n)$. We also set $ t_{\max} \ll n$, so to activate the $j^{th}$ edge, the algorithm needs to compute at most $O\big((n + j t_{\max}) N s_{\max}^2\big)$ sufficient statistics tables. In the case of edge grafting, the algorithm needs to compute $O(n^2 N s_{\max}^2)$ to start grafting the first edge. Since $(n + j t_{\max}) \ll n^2$, our method provides a drastic speedup, as we circumvent the quadratic term in $n$ and replace it with a linear term. 
Finally, between each activation step, we must compute approximate expectations to obtain the gradient. Approximate inference techniques typically take time linear in the number of active edges.
\cref{table:time} summarizes the different time complexity for the discussed algorithms.

\section{Experimental Evaluation}

Our experiments include simulation of reservoir-based search and evaluation of the speed and quality of learning from synthetic and real data.
We use datasets of different variable dimensions and dataset sizes. In the synthetic setting, we simulate MRFs and generate data using a Gibbs sampler. In the real setting, we use two datasets. We use exhaustive edge grafting (labeled ``EG'' in figures) as our primary baseline. To show the benefits of the reservoir, we also include best-choice edge grafting with no reservoir (``first hit''), which activates the first edge that passes the edge-activation condition. 

\subsection{Benefits of the Reservoir}

We analyze the increase in edge quality that the reservoir provides. The reservoir management protocol mimics a reservoir of randomly selected entries from the full population. Asymptotically, the relative percentile of any randomly chosen edge is equivalent to a uniform random variable in the range $[0, 1]$. The rank of the best edge in a size-$|R|$ reservoir is thus analogous to the minimum of $|R|$ uniform draws. It is well known that the expected value of this minimum is $1/(|R| + 1)$.  We simulate the behavior in finite settings, sampling $|R|$ ranks from the list of all possible numbers from 1 to $n \choose 2$ and taking the minimum. We then plot the average minimum rank over 100 trials, using $n=400$ with values of $|R|$ from 1 to 500. \cref{fig:reservoir} plots the average ranks over the trials. The results suggest that a small reservoir provides significant gains over using first hit ($|R| = 1$). Using an unlimited reservoir---i.e., grafting---only provides negligible gains over a small reservoir. Since grafting is equivalent to a reservoir of size $400 \choose 2$, or 79,800, the benefits of using a reservoir are evident. 

\FloatBarrier

\begin{figure*}[b]
\centering
%\subfigure[50 nodes and 150 edges]{
%    \includegraphics[width=0.3\textwidth]{figures/50OBJ.eps}
%}
%~
\subfigure[200 nodes and 600 edges]{
    \includegraphics[width=0.31\textwidth]{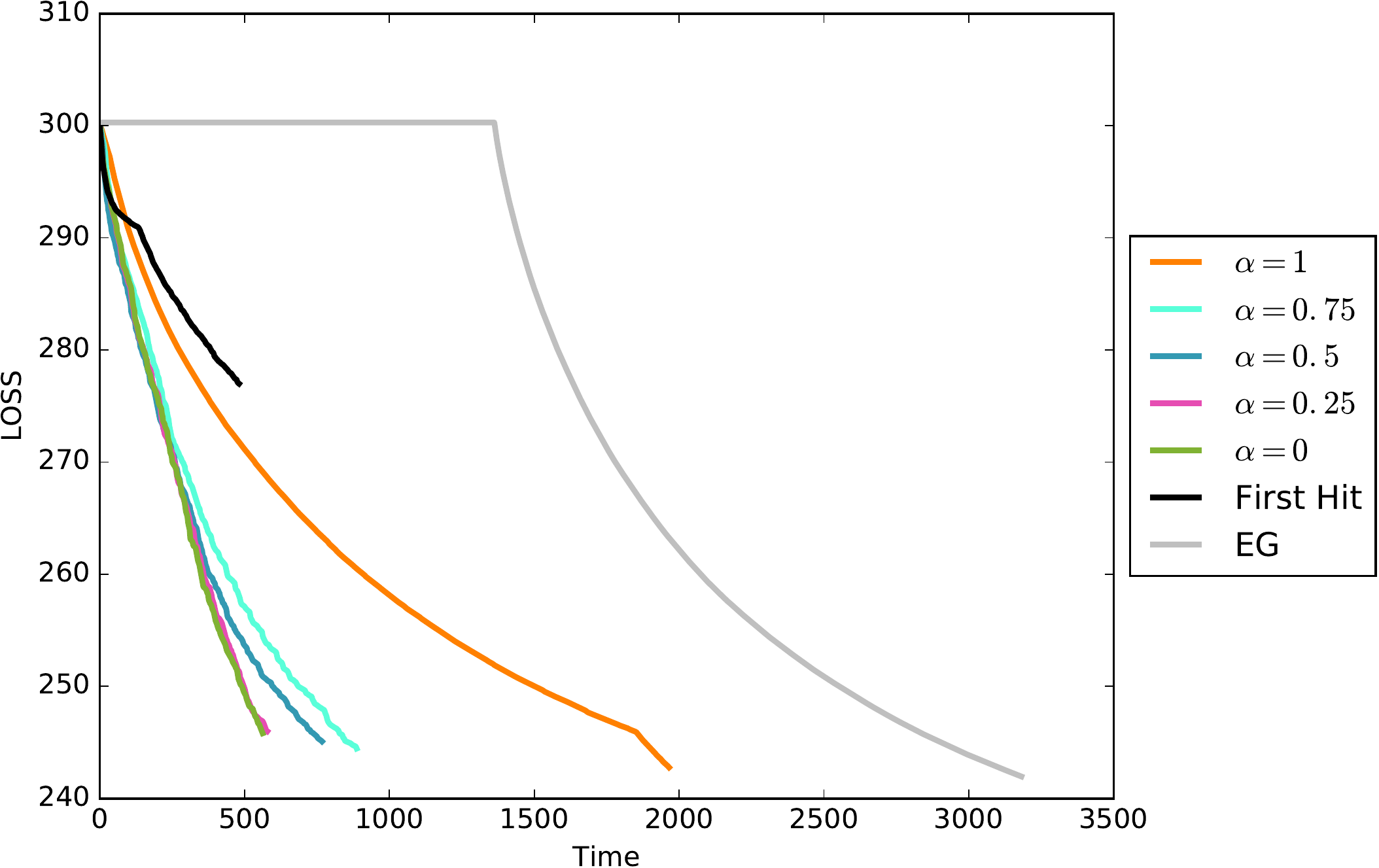}
}
\subfigure[400 nodes and 1,200 edges]{
    \includegraphics[width=0.31\textwidth]{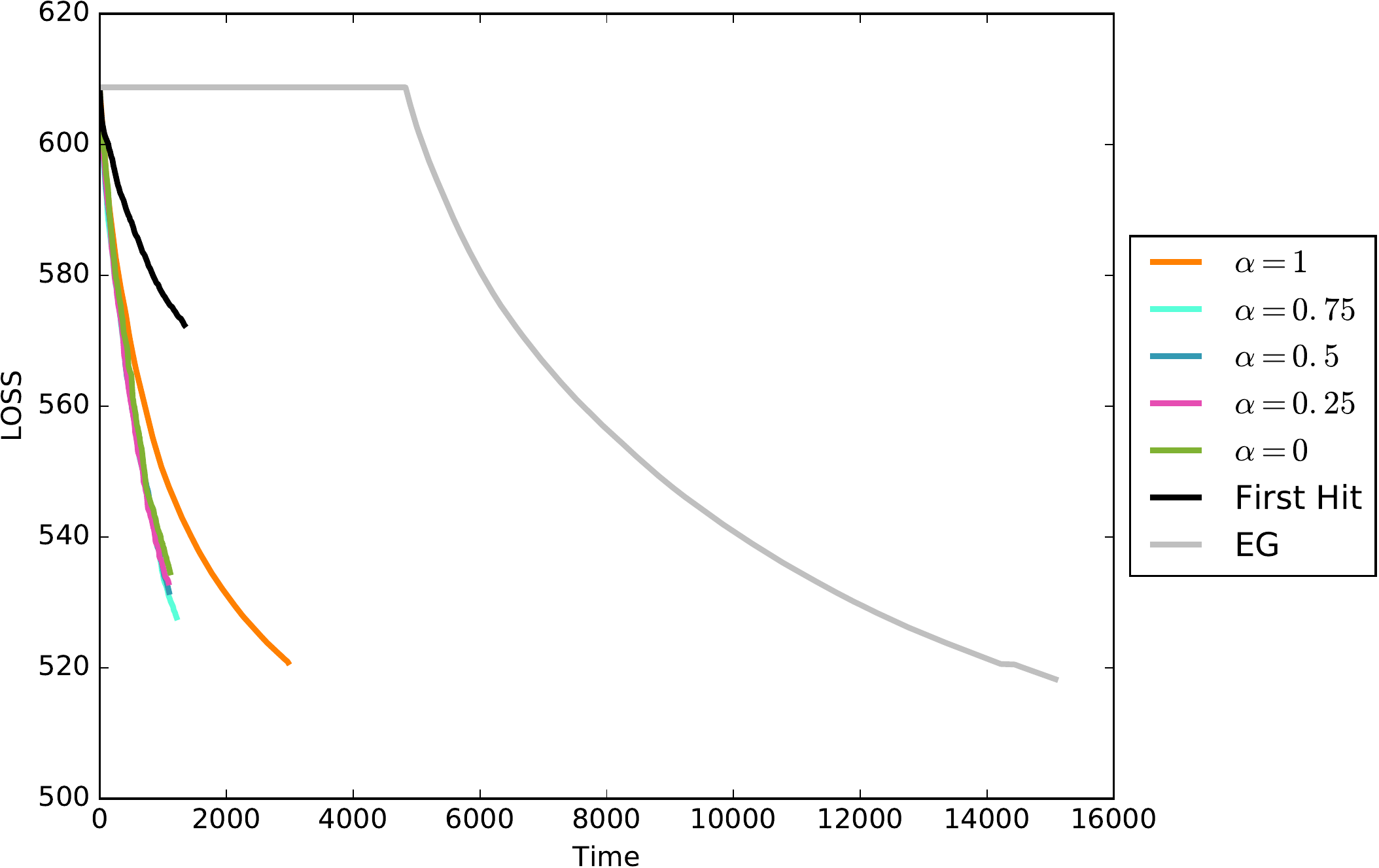}
}
\subfigure[600 nodes and 1,800 edges]{
    \includegraphics[width=0.31\textwidth]{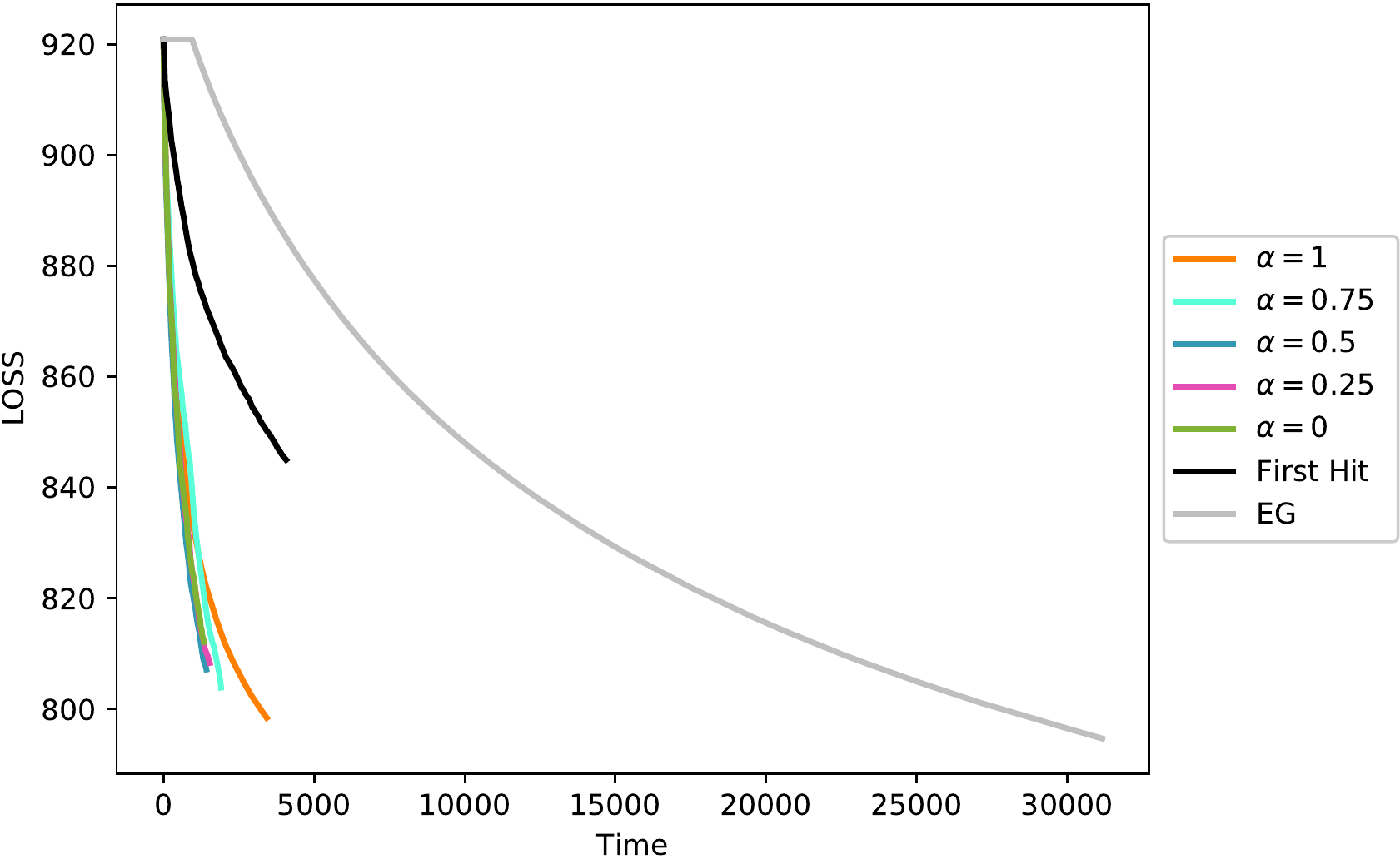}
}
\caption{Loss vs.~time (seconds) for varying MRF sizes with $O(n)$ edges.}
\label{fig:obj}
\end{figure*}

\begin{figure*}[tb]
\centering
\subfigure[200 nodes and 600 edges]{ 
   \includegraphics[width=0.31\textwidth]{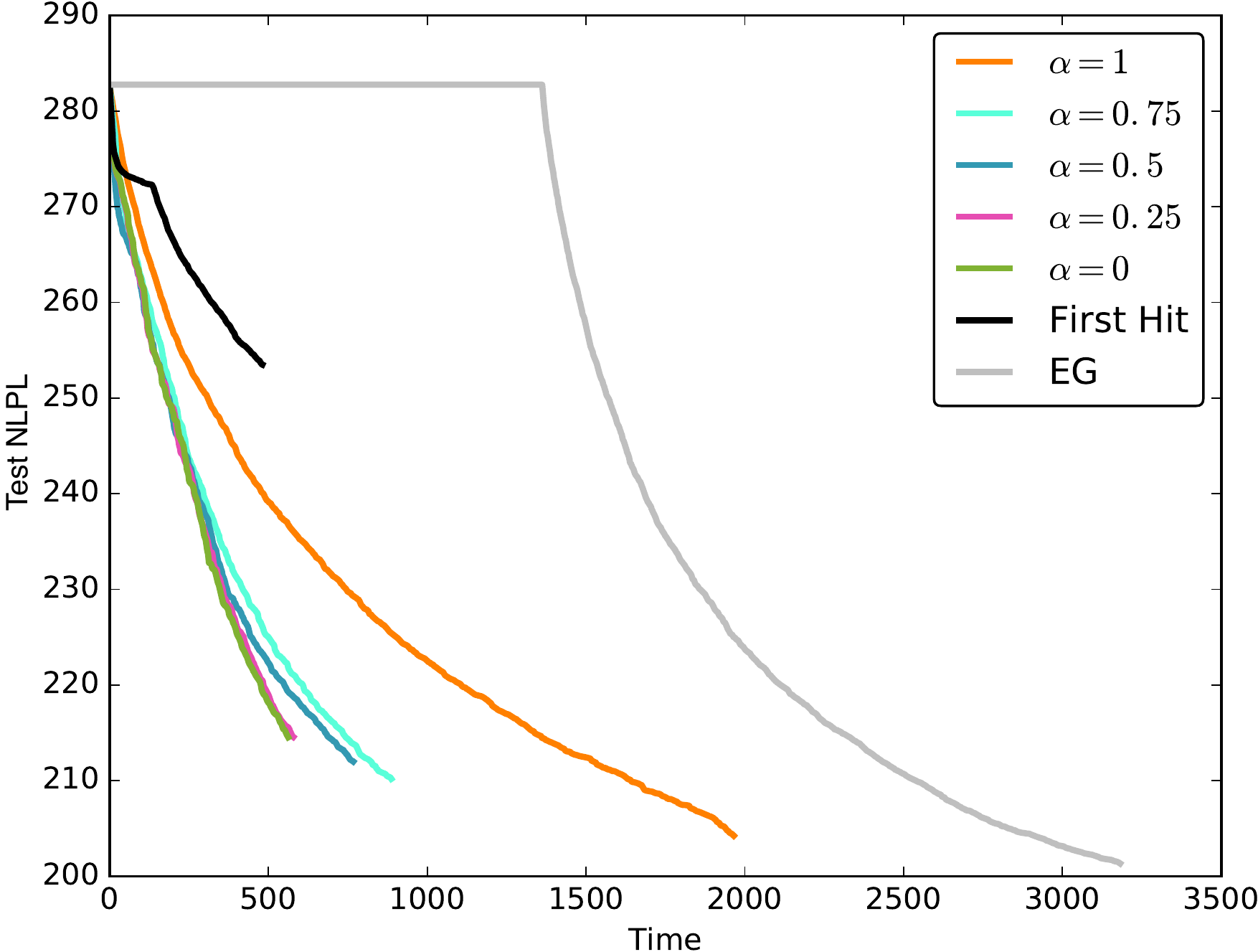}
}
\subfigure[400 nodes and 1,200 edges]{
   \includegraphics[width=0.31\textwidth]{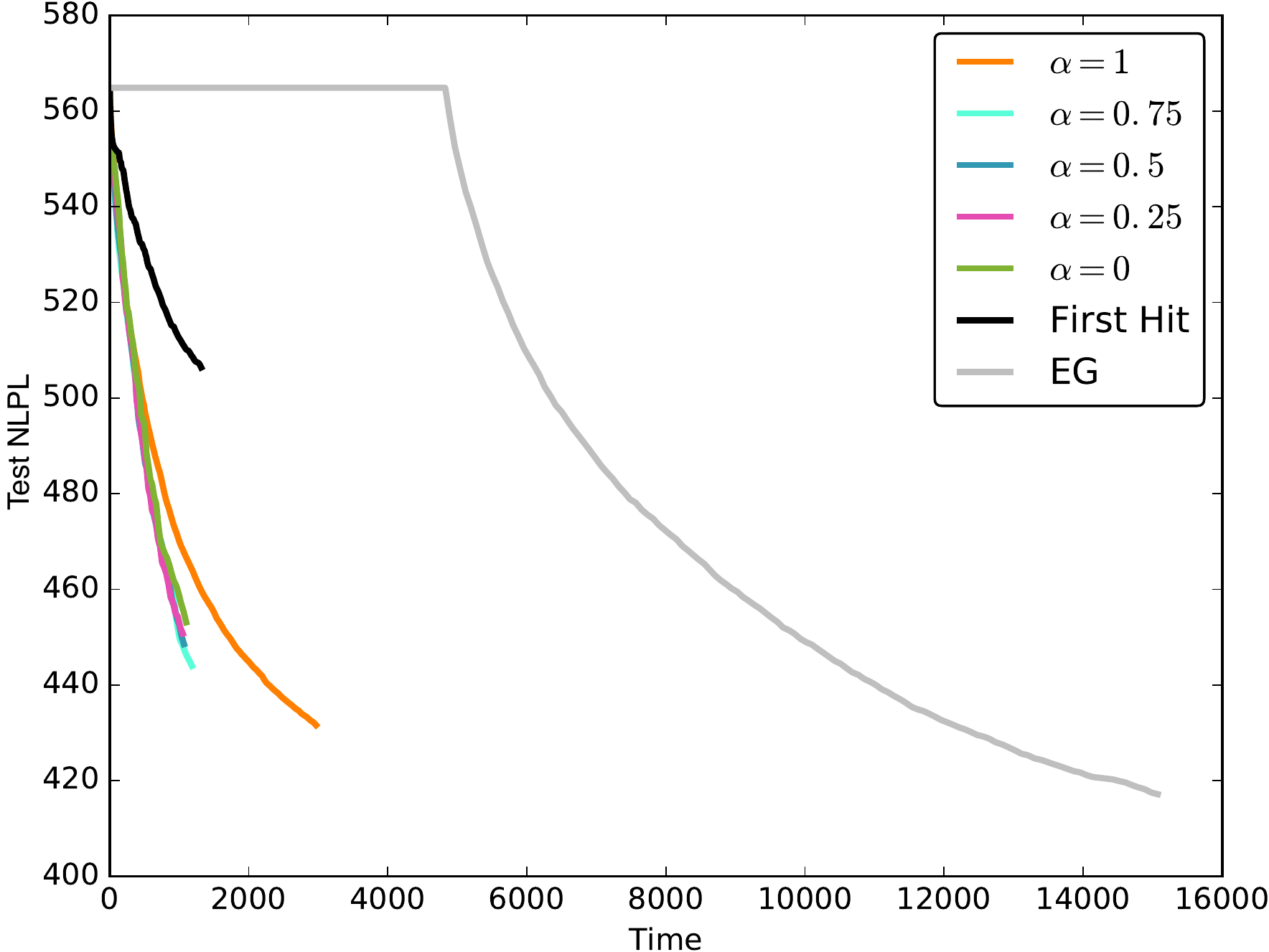}
}
\subfigure[600 nodes and 1,800 edges]{
   \includegraphics[width=0.31\textwidth]{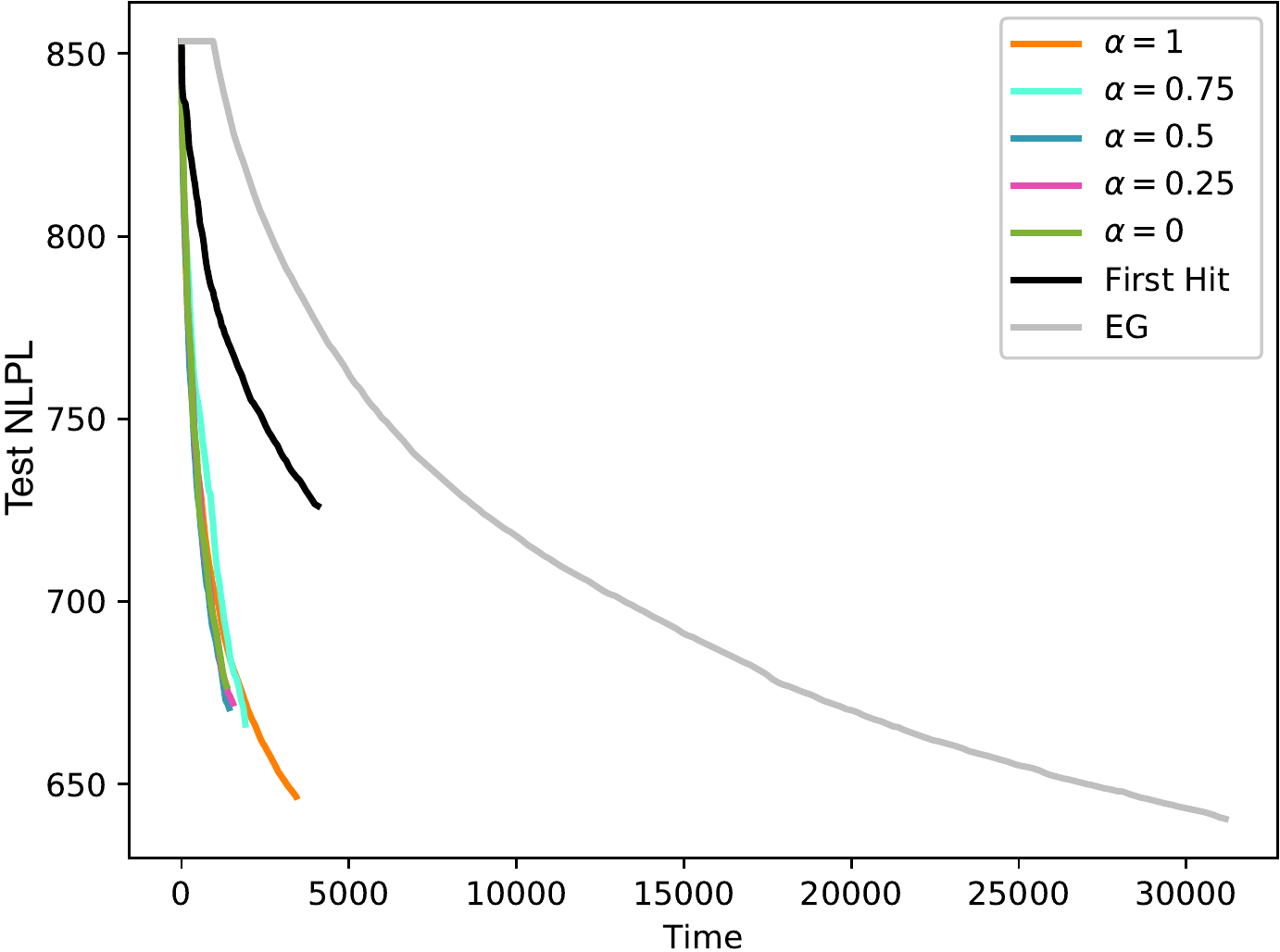}
}
\caption{Negative log pseudo-likelihood (NLPL) vs.~time (seconds) for varying MRFs sizes with $O(n)$ edges.}
\label{fig:nll}
\end{figure*}

%We measure how well the learned MRF models the data distribution and how well it recovers the true underlying graphical model structure. We record the learning objective to compare how well each method optimizes. And we measure how well the learned MRFs represent the true data distribution with the pseudo-likelihood \cite{besag1977efficiency} of held-out test data.
%As a structure-recovery metric, we use recall, defined as the fraction of true edges recovered.
%We monitor recall over time. The slope of recall against time is an indicator of precision, as a high slope indicates that a method adds relevant edges, and a small slope indicates the activation of several false-positive edges. 

\subsection{Synthetic Data}

We construct random, scale-free structured MRFs with variables having five states each. We generate preferential-attachment graphs \cite{albert2002statistical} of size 200, 400, and 600, with 498,500, 1,997,000, and 4,495,500 parameters, respectively.
We then use Gibbs sampling to generate a set of $20,000$ data points, which we randomly split as training and held-out test data. We use grid search to tune the learning parameters. We fix $|R| = n$, and $t_{\max} \ll n$. 
See the appendix for more details on the experimental setup.
%Our experiments evaluate the the quality of learned MRFs during learning and also the impact of reservoir-threshold parameter $\alpha$ on the learning curve.

\begin{wrapfigure}{O}{0.5\textwidth}
\centering
    \includegraphics[height=1.75in]{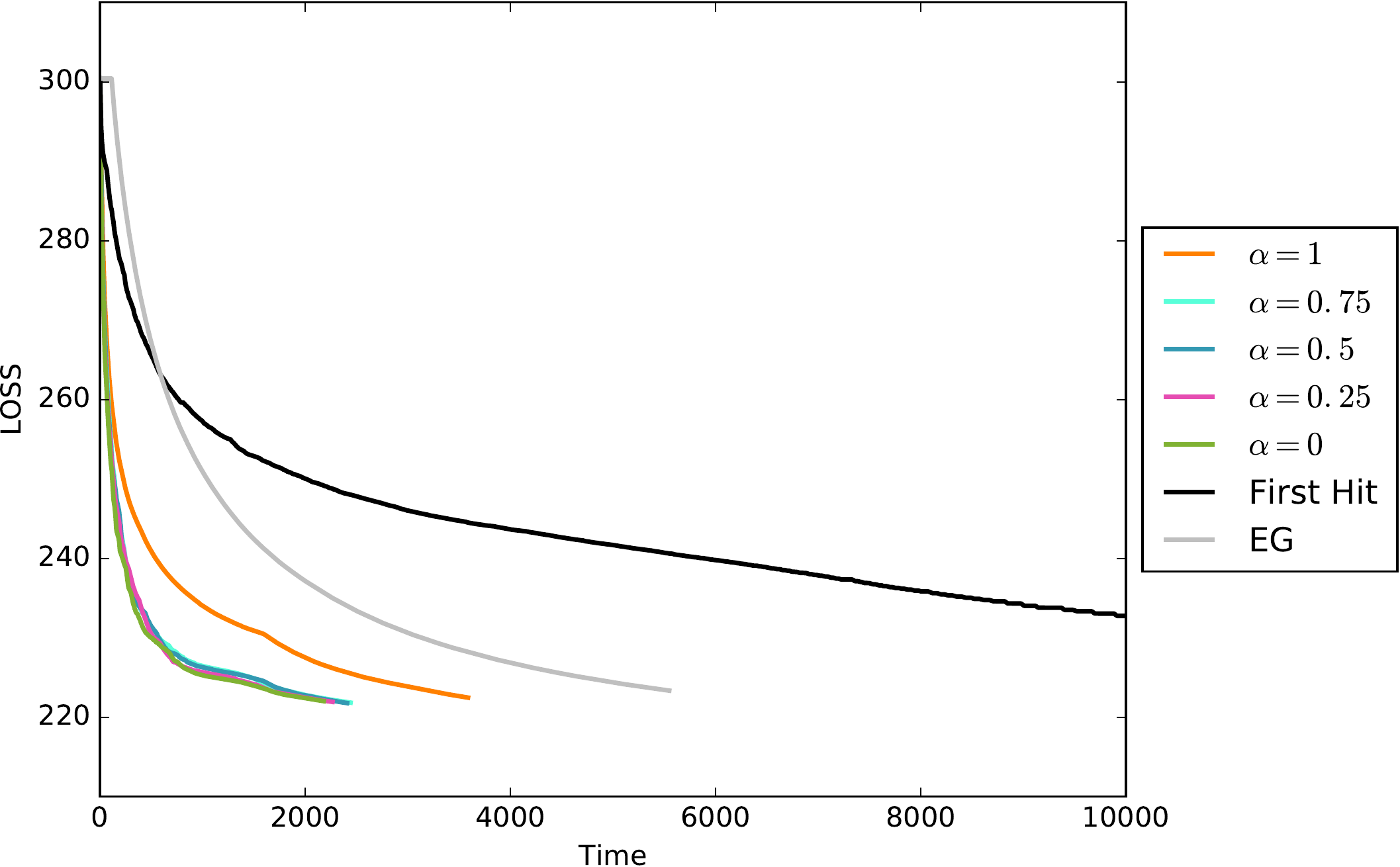}
\caption{Objective values vs.~seconds during full convergence of all methods (200 nodes). The first-hit method takes around 30,000 seconds to converge, so we truncate the horizontal axis for a better view of the other methods.}
\label{fig:convergence}
\end{wrapfigure}

\textbf{Faster convergence} ~~ We first investigate the behavior of different methods when executed until no violating edge remains in the search space. We measure the different methods' convergence speeds and confirm that they lead to similar solutions. \Cref{fig:convergence} shows that all methods reach a similar solution, but best-choice edge grafting has a faster convergence rate. The first-hit baseline starts with a fast descent rate, but its activation of lower-violation edges eventually causes it to converge slower than edge grafting. Using a reservoir maintains a steep descent until convergence. These experiments also confirm that edge grafting suffers a major cost of computing sufficient statistics, causing it to start optimization after best-choice edge grafting has nearly converged.

\begin{figure*}[tb]
\centering
\subfigure[200 nodes and 600 edges]{
   \includegraphics[width=0.31\textwidth]{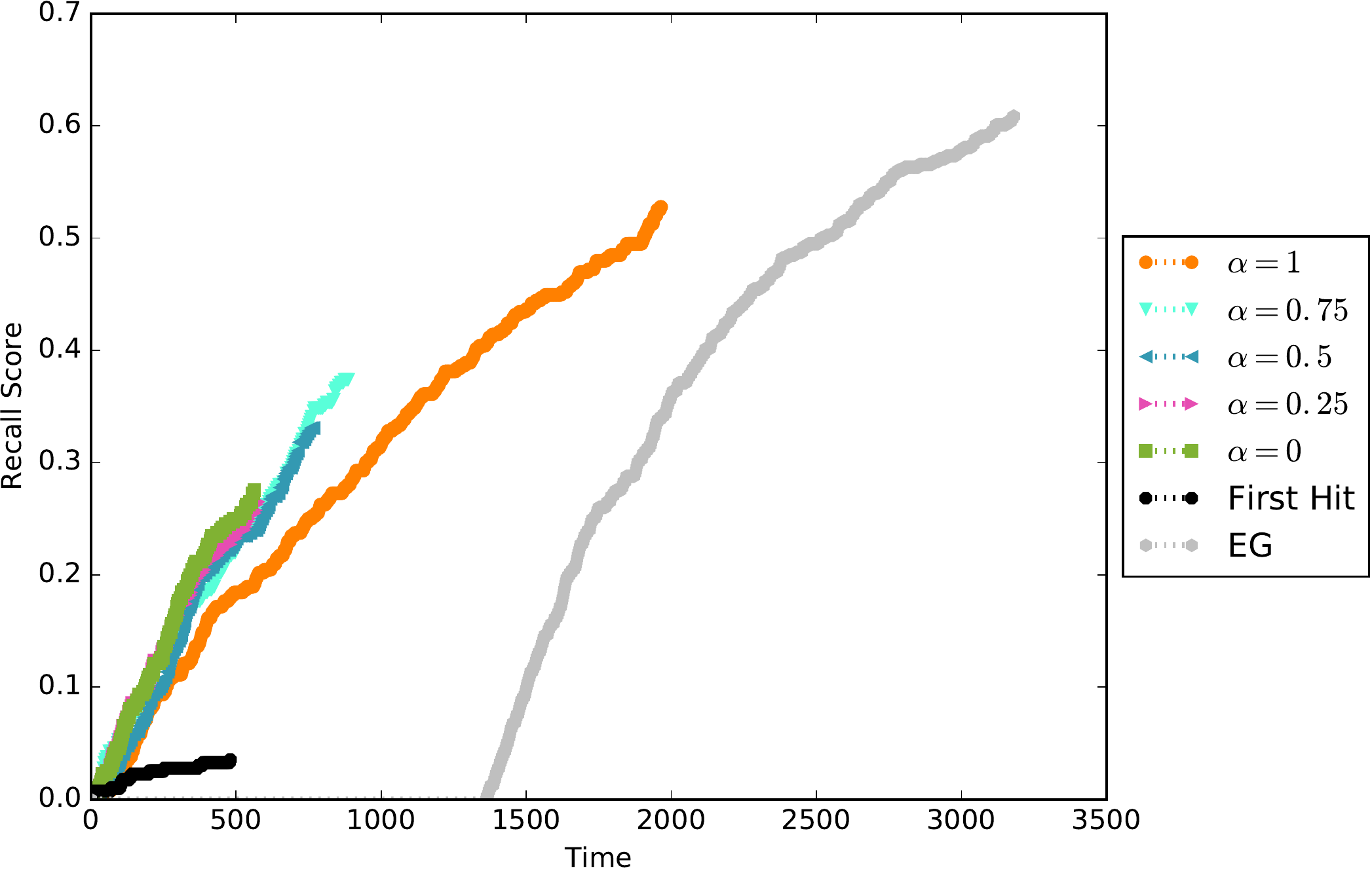}
}
\subfigure[400 nodes and 1,200 edges]{
   \includegraphics[width=0.31\textwidth]{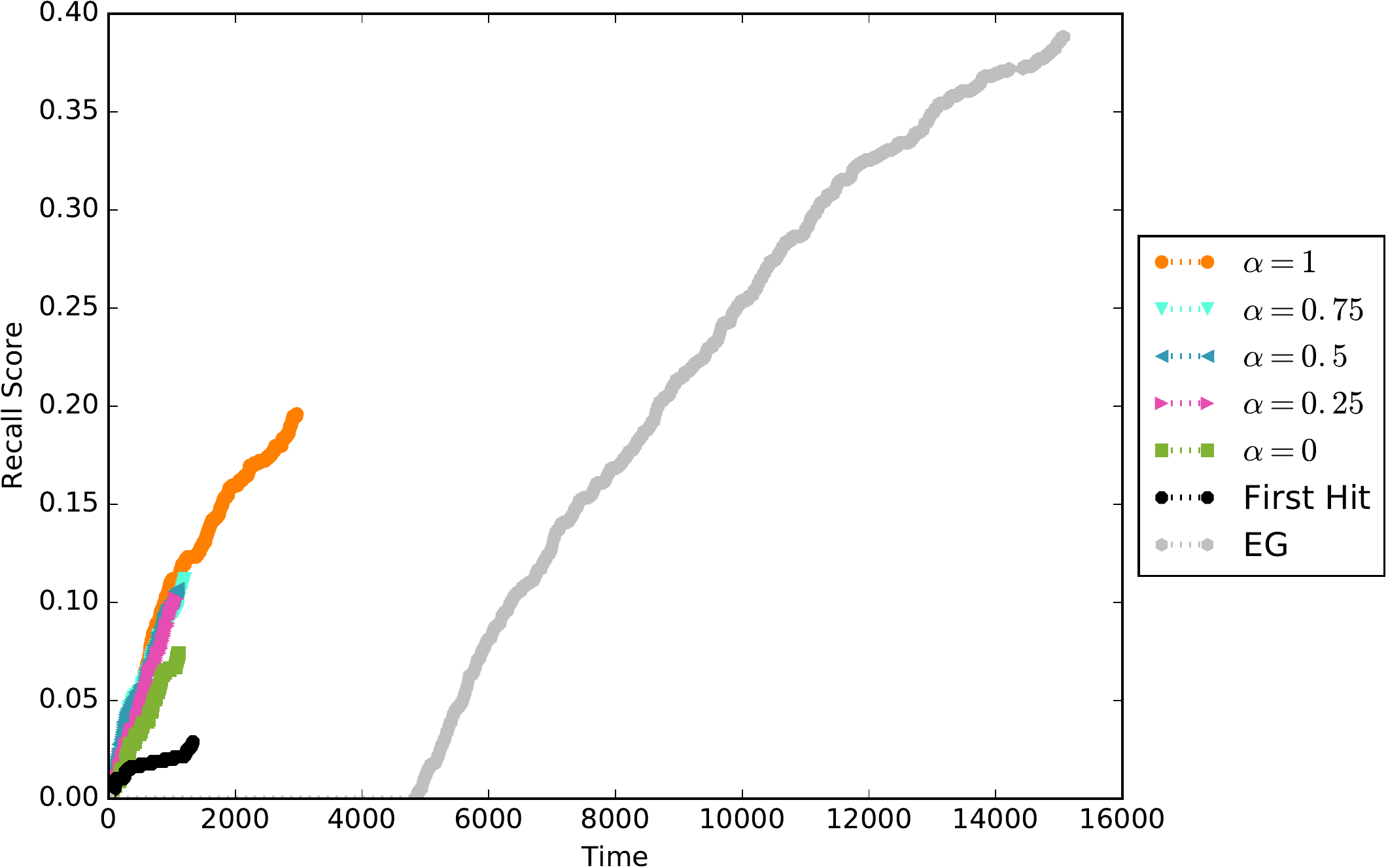}
}
\subfigure[600 nodes and 1,800 edges]{
   \includegraphics[width=0.31\textwidth]{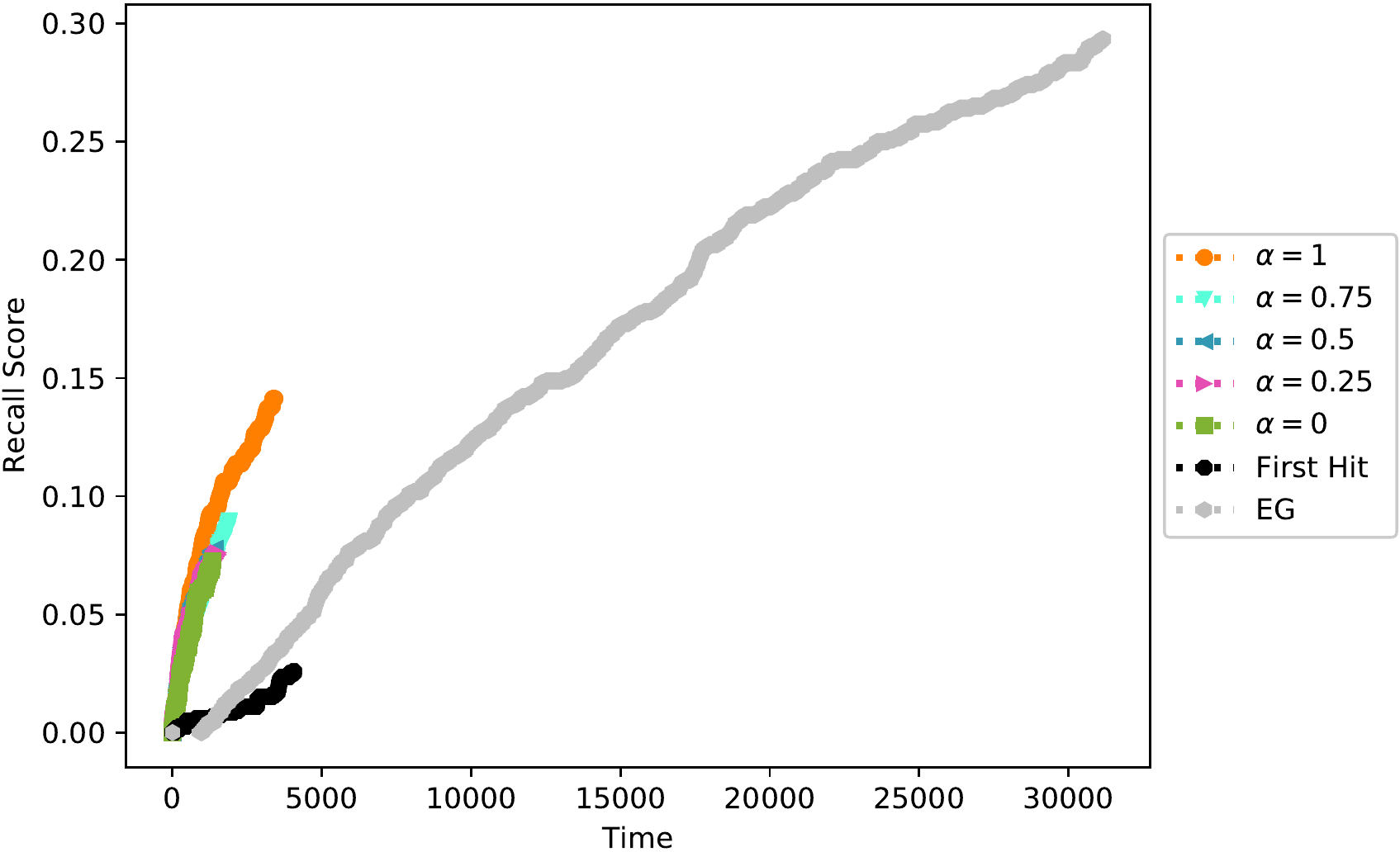}
}
\caption{Recall vs.~time (seconds) for varying MRFs sizes with $O(n)$ edges.}
\label{fig:recall}
\end{figure*}

\textbf{Controllable tradeoff between learning speed and quality} ~~ In subsequent experiments, we fix the maximum number of activated edges to $3n$, stopping early when each algorithm exhausts this limit. We first measure the objective value during learning and plot it over running time in \cref{fig:obj}. We observe similar trends: Best-choice edge grafting provides significant speedups over exhaustive and first-hit edge grafting. Higher $\alpha$ values enable better quality but result in a slower optimization. Experiments on held-out testing data (\cref{fig:nll}) show that there is a positive correlation between the learning objective and the test negative log pseudo-likelihood (NLPL), which confirms that the learned models do not over-fit even with small values of $\alpha$. The first-hit baseline reaches the edge limit faster, but its lower-quality edges cause slower convergence (see \cref{fig:convergence}) and a lower quality model (\cref{fig:obj,fig:recall}).

\textbf{Faster edge activation}~~ 
For increasing amounts of variables, the learning gap between best-choice edge grafting and exhaustive edge grafting increases, which demonstrates the better scalability of the best-choice algorithm.
In the smaller graphs, exhaustive edge grafting has the advantage that its greedy search for the worst-violating edge enables large improvements in the objective (and pseudo-likelihood). However, in the larger graphs, even though edge grafting precomputes sufficient statistics, the remaining $O(n^2)$ cost of the greedy search causes its objective to descend at a slower rate than the best choice variants, which avoid this exhaustive search. This high cost is especially evident in the 600-variable problems (e.g., \cref{fig:obj}c), where it takes nearly ten times as long for edge grafting to add the desired number of edges as the reservoir-based best-choice approaches.

\begin{wrapfigure}[16]{O}{0.5\textwidth}
\centering
\includegraphics[height=1.5in]{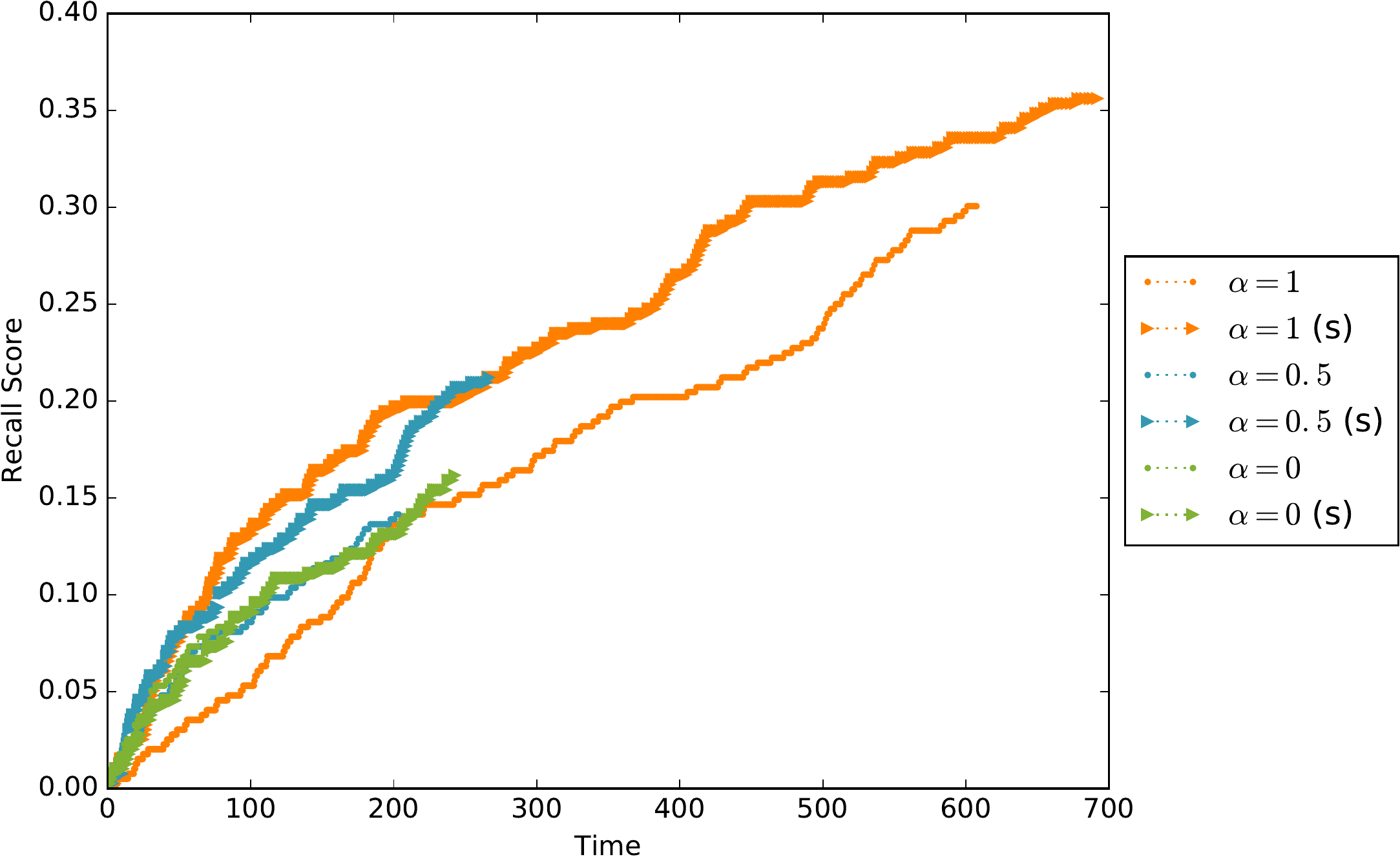}
\caption{Role of structure heuristics in improving the quality of the learned MRF over training time (seconds). The proposed heuristics (labeled ``s'') produce higher recall for an MRF of 200 nodes and 600 edges.}
\label{fig:sheuristics}
\end{wrapfigure}

\textbf{Structure learning quality}~~ \cref{fig:recall} plots the recall of true edges over time until the maximum number of added edges is met. Increasing values of $\alpha$ lead to better recall for different MRF sizes. However, this comes at the cost of a more expensive learning optimization. In fact, for smaller values of $\alpha$, best-choice edge grafting tends to activate more edges but with lower quality, which introduces a greater number of false-positive edges. 
This result suggests that, while higher values of $\alpha$ help recover the true structure, if the goal of learning the MRF structure is to produce a good generative model, then lower $\alpha$ values speed up learning with only a small loss in quality.

\textbf{Effectiveness of structure heuristics}~~ To evaluate the effectiveness of structure heuristics, \cref{fig:sheuristics} plots the edge recall---for the same limited number of activated edges---with and without the structure-based priority queue reorganization. The significant gap between the recall curves shows that the heuristics produce a higher-quality edge stream from the priority queue, resulting in a higher-quality reservoir and more relevant edges to activate.

%Figure \ref{fig:NLPL} shows the evolution of NLPL, figure \ref{fig:F1} shows the evolution of F1 scores, figure \ref{fig:Loss} shows the evolution of the learning objective.

%To showcase the benefits of search space reorganization we plot the evolution of the priority queue of priority edge grafting. The figure shows that as learning progresses, relevant edges get pushed to the head of the queue. This means that when we start next search iteration we spend less activation iterations.

%On the other hand we compare the would like to see how priority edge grafting reduces the number of sufficient statistics tables. To see the effect of the search space reorganization we compare priority edge grafting to queue edge grafting. Queue edge grafting is a version of priority edge grafting where priority reassignment is prohibited. We can see that priority reassignment yields the computation of lesser number of sufficient stats tables.

\subsection{Real Data}

We use the Jester joke ratings \cite{goldberg2001eigentaste} and Yummly recipes \cite{yummly} datasets.
Jester is a joke ratings dataset containing $100$ jokes and $73,421$ user ratings. We use jokes as variables and user ratings as instances. We map the ratings interval from a continuous interval in $[0, 20]$ to a discrete interval from $1$ to $5$, leading to $124,250$ parameters.
We sample $10,000$ recipes from the Yummly data, using the 489 most common ingredients (which occur in at least 150 recipes) as binary variables and each recipe as a data instance, leading to $478,242$ parameters.
%Our aim with these experiments is to assess how the proposed framework performs on different and diverse real datasets. 
%As in the synthetic setting, we monitor the learned MRFs over time and evaluate them using their learning objectives and their negative log pseudo-likelihood.

\begin{figure*}[h!]
\centering
\subfigure[Jester Objective]{
    \includegraphics[width=0.31\textwidth]{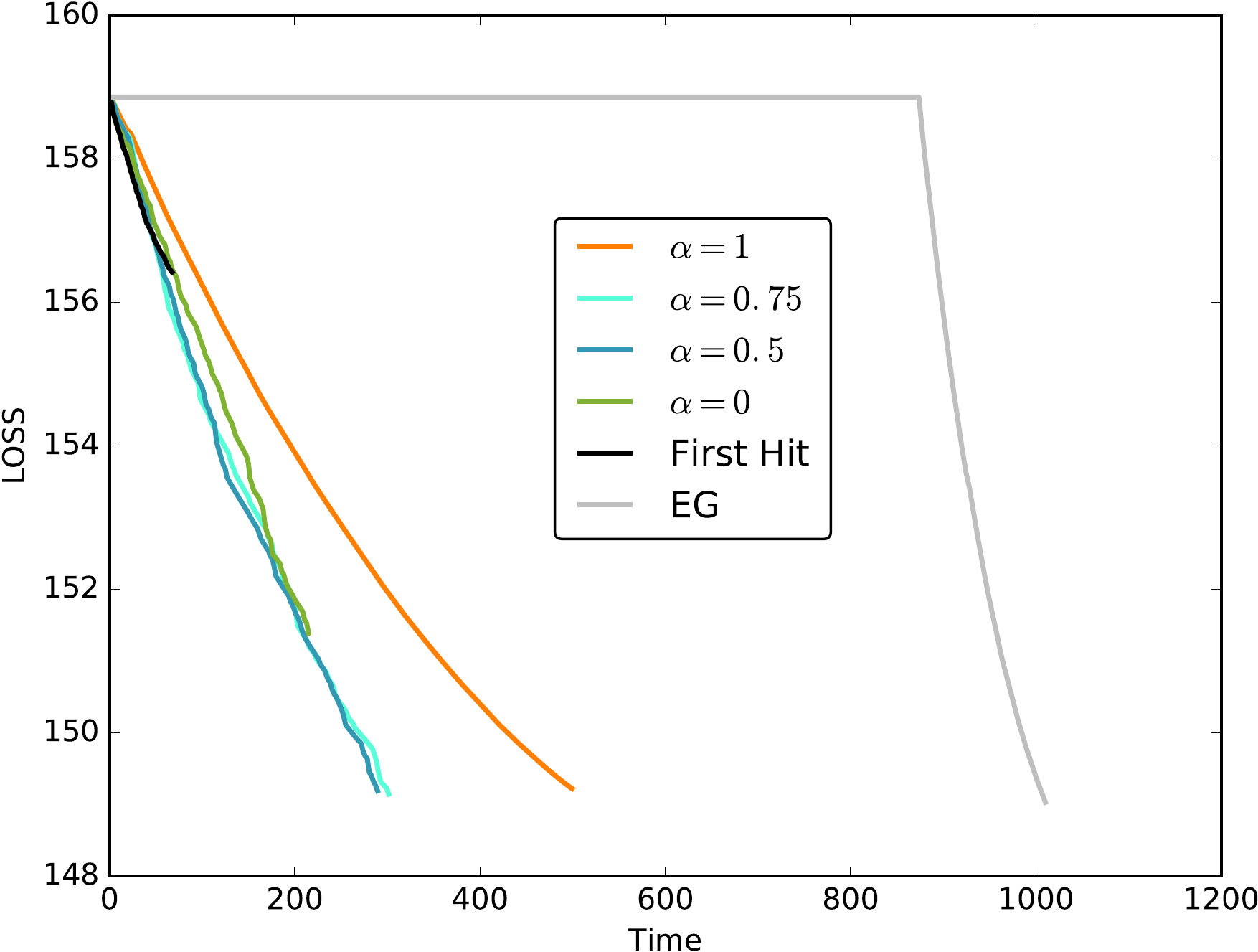}
}
~~~~~~~~~~~~
\subfigure[Jester NLPL]{
    \includegraphics[width=0.31\textwidth]{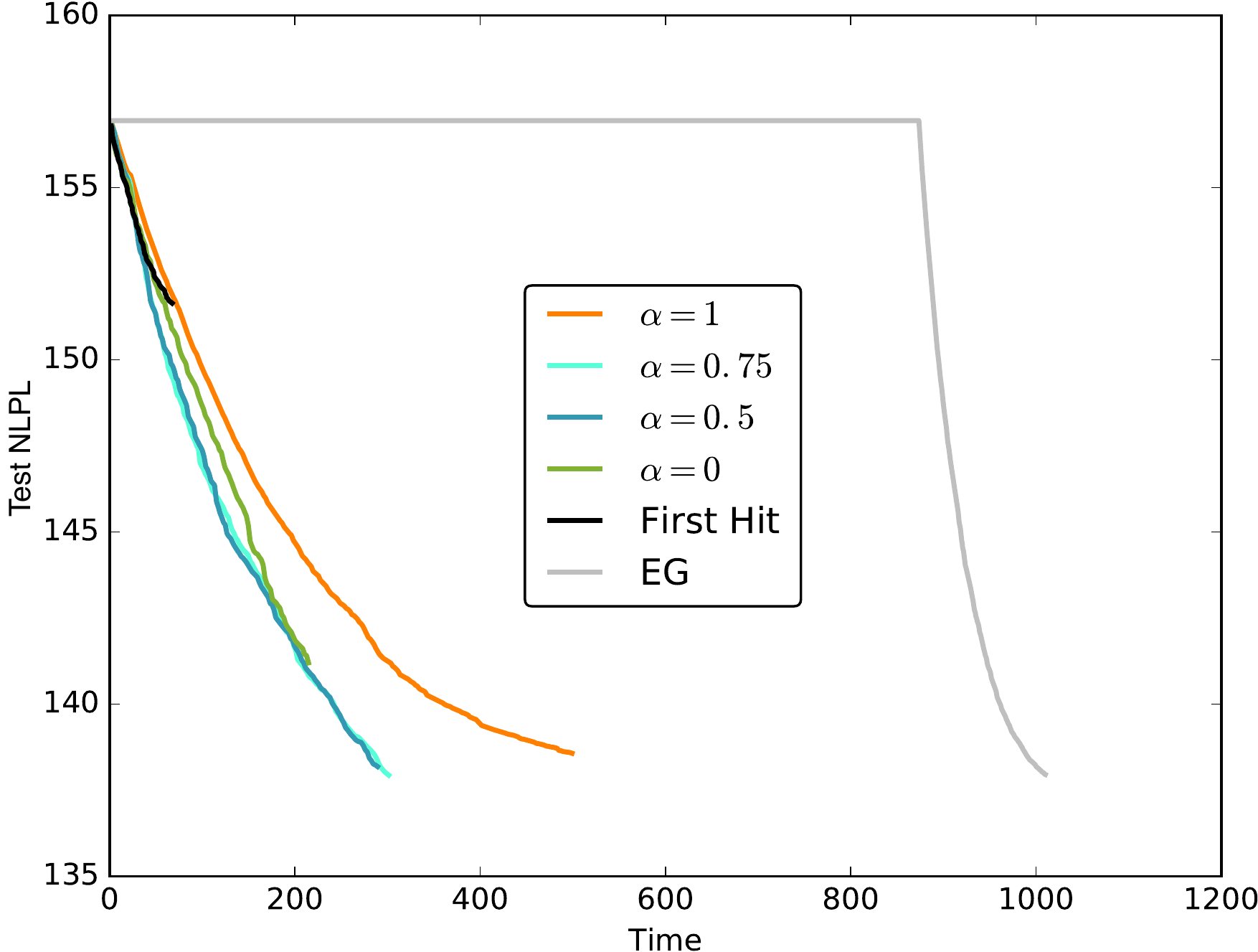}
}
\\
\subfigure[Yummly Objective]{
    \includegraphics[width=0.31\textwidth]{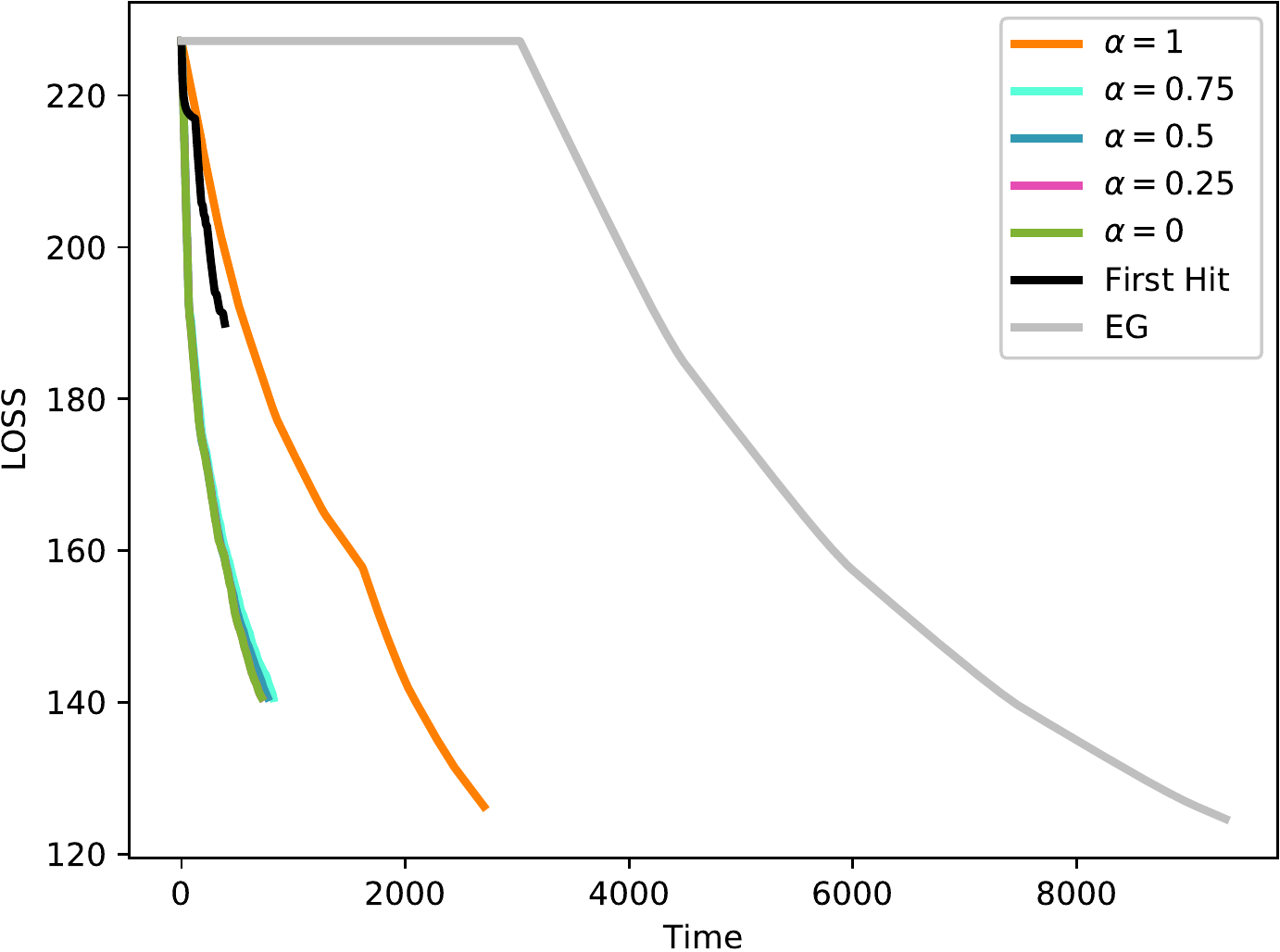}
}
~~~~~~~~~~~~
\subfigure[Yummly NLPL]{
    \includegraphics[width=0.31\textwidth]{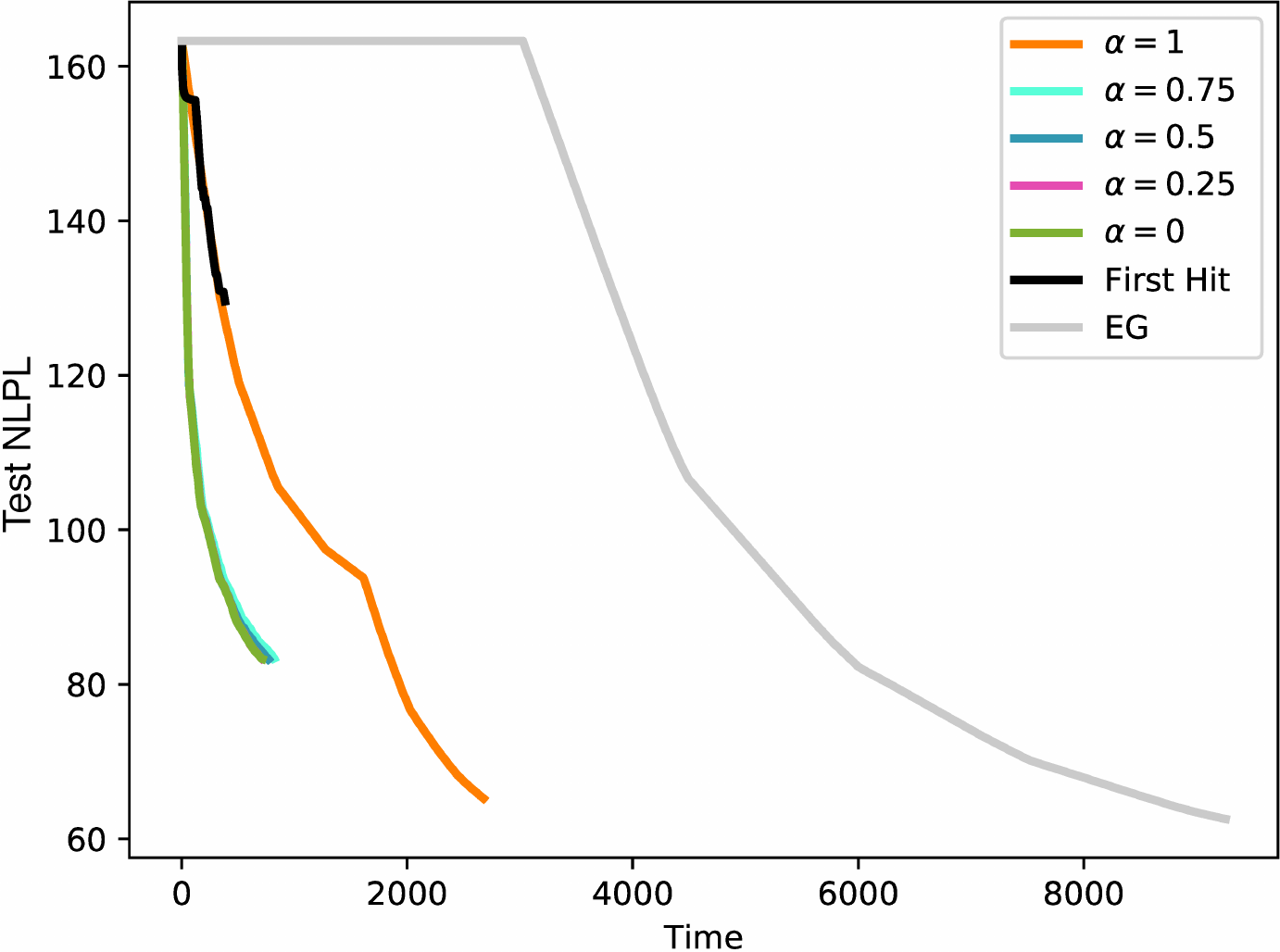}
}
\caption{Learning objective (a, c) and negative log pseudo-likelihood (b, d) vs.~seconds for Jester and Yummly.}
\label{fig:real}
\end{figure*}

\Cref{fig:real} shows a positive correlation between the minimization of the learning objective and the testing negative log pseudo-likelihood. The results are similar to those from the synthetic experiments, where best-choice edge grafting converges faster than edge grafting, and smaller values of $\alpha$ result in a faster convergence. 
%In datasets where there are low dependencies between variables, such as in the case of USDA plants dataset, the difference in the quality for different values of $\alpha$ becomes more contrasted. Hence, it is recommended to set a high value of $\alpha$. 
%The first-hit baseline displays a behavior similar to that observed in the synthetic setting, as it converges quickly but produces low quality MRFs compared to the rest of the methods. 
The different datasets illustrate the higher scalability of best-choice edge grafting, as its advantage in convergence time over edge grafting increases with dimensionality.

\section{Conclusion} 

We presented best-choice edge grafting, a method based on a group-$\ell_1$ formulation and reservoir sampling. This incremental method activates edges instead of features and uses a reservoir to approximate greedy activation. Theoretical analysis of the iteration complexity shows that the method provides higher scalability than exhaustive edge grafting by avoiding its major bottlenecks. Experiments on synthetic and real data demonstrate that best-choice edge grafting yields faster convergence on multiple datasets, while also achieving structure recovery and predictive ability similar to the more costly exhaustive edge grafting.

%\FloatBarrier

%\section*{References}

\bibliographystyle{abbrv}
\bibliography{biblio}

\appendix

\section{Edge Grafting and Best-Choice Edge Grafting Algorithms}

We summarize edge grafting, as discussed in Section 3.1, in Algorithm \ref{alg:edgeGrafting}.

\begin{algorithm*}[h!]
\begin{algorithmic}[1]
\STATE Define EdgeNum as the maximum allowed number of edges to graft
\STATE Initialize $E=\emptyset$ and $ F = \text{ set of all possible edges}$
\STATE Compute sufficient statistics of $e$ $\forall e \in F$ \COMMENT{cost: $O(n^2 N s_{\max}^2)$} \label{op0}
\STATE AddedEges $ = $ 0
\STATE Continue $=$ True
\WHILE{EdgeNum $>$ AddedEges and Continue}
\STATE Compute score $s_e$ $ \forall e \in F$ \COMMENT{cost: $O(n^2 s_{\max}^2)$} \label{op1}
\STATE $e^* = \arg\max_{s \in F} s_e$. \COMMENT{cost: $O(n^2$)} \label{op2}
\IF{$s_{e^*} > \lambda$}
\STATE $E = E \cup \{e^*\}$; $F = F \backslash \{e^*\}$
\STATE AddedEges $ = $ AddedEges $ + 1$
\STATE Perform optimization over new active set
\ELSE
\STATE Continue = False
\ENDIF
\ENDWHILE
\end{algorithmic}
\caption{Edge Grafting}
\label{alg:edgeGrafting}
\end{algorithm*}

\begin{figure*} [h!]
       \centering
       \caption{High-level operational scheme of the edge activation mechanism.}
       \begin{tikzpicture}
       \tikzstyle{connect}=[-latex, thick, auto, main node/.style={circle,draw, inner sep=1pt,minimum size=.75cm}, baseline= (3.south)]
         \node[rectangle split, rectangle split parts=6,
        draw, rectangle split horizontal,text height=.5cm,text depth=0.5cm,inner ysep=0pt] (rr) {};
        
         \node[draw,diamond,minimum size=1.5cm, right=1cm of rr](c2){$C_2?$};
         \draw[->] (rr.east) -- +(20pt,0) node[above] {$e$};
         \draw[->] (c2.east) -- +(15pt,0) node[above] {yes};
         \draw[->] (c2.south) -- +(0pt,-0.75) node[above, left] {no};
         % the rectangular shape with vertical lines
        \node[rectangle split, rectangle split parts=6,
        draw, rectangle split horizontal,text height=.5cm,text depth=0.5cm,inner ysep=0pt, right=1cm of c2] (wa) {};
        
         \node[draw,diamond,minimum size=1.5cm, right=0cm of wa](deadline){$t_{\max}?$};
         \draw[->] (deadline.east) -- +(25pt,0) node[above, black] {activate};
        
         \fill[white] ([xshift=-\pgflinewidth,yshift=-\pgflinewidth]wa.north west) rectangle ([xshift=-15pt,yshift=\pgflinewidth]wa.south);

        \node[rectangle split, rectangle split parts=6,
        draw, rectangle split horizontal,text height=.5cm, text depth=0.5cm,inner ysep=0pt,  below=1cm of c2] (ll) {};

         % the arrows and labels
         \node[]  (rrn) [left=.001cm of rr] {$pq$};
         \node[]  (blankr) [left=0.0cm of wa] {};
         \node[]  (rn) [right=0.15cm of blankr] {$R$};
         \node[align=center,below = .5cm of ll.south] at (ll.south) {$L$};
        %\node[align=center,below] at (se2.south) {Server \\ process};

        \path[every node/.style={font=\sffamily\small}]
         (rr) edge[loop above] node {reorganize} ()
         (wa) edge[loop above] node {update} ()
         (ll) edge [connect, dashed, gray] node {refill(L)} (rr.south)
         (wa.south) edge [connect, dashed, gray] node {$R_{\min}$} (ll)
         ;

       \end{tikzpicture}
             \label{fig:reorganization_mechanism}
       \end{figure*}
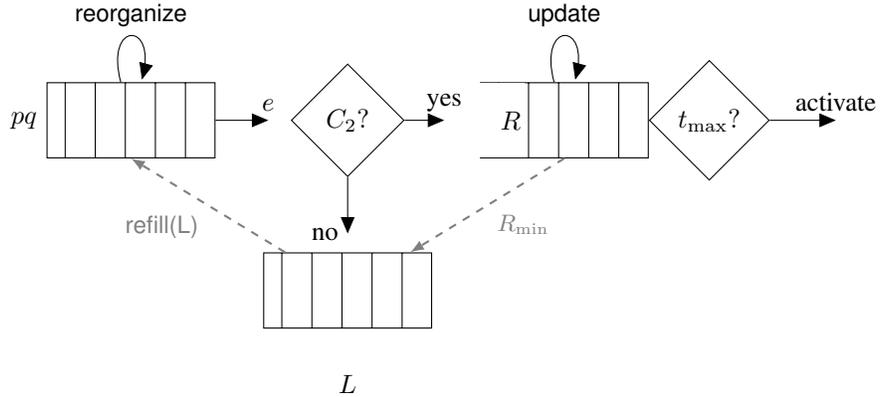

Figure \ref{fig:reorganization_mechanism} presents a high-level description of the best-choice edge-activation mechanism:  From left to right, the first structure is the priority queue ($pq$) initialized with the set of all possible edges. The next diamond-shaped box represents the activation test $C_2$, after which an edge is either added to the reservoir ($R$) or to the frozen edge container $L$. The $t_{\max}$ box represents when the maximum number of edge tests is reached, after which edges are activated. The gray dashed line on the right ($R_{\min}$) indicates the injection of the minimum scoring edge in $R$ into $L$, when $R$ is full. The gray dashed line on the left (refill(L)) indicates refilling the priority queue with the frozen edges once it is emptied.
     
The priority-queue reorganization subroutine is summarized in Algorithm \ref{alg:reorganizepq}, and the activation mechanism is presented in Algorithm \ref{alg:Activation}. These form the underlying components to construct the main best-choice edge grafting framework in Algorithm \ref{alg:OEG}.

\begin{algorithm*}[h!]
\begin{algorithmic}[1]
\STATE Compute centrality measures over partially constructed MRF graph $G$
\STATE Construct the hub set $H$ \COMMENT{cost: $O(n)$}
\FOR{$h \in H$} 
\FOR{$n \in V$}
\STATE $pq[(h,n)] = pq[(h,n)] - 1$ \COMMENT{total loop cost: $O(|H| n \log(n))$}
\ENDFOR
\ENDFOR
\end{algorithmic}
\caption{Reorganize PQ}
\label{alg:reorganizepq}
\label{alg:PEG}
\end{algorithm*}

\begin{algorithm*}[h!]
\begin{algorithmic}[1]
\STATE Reorganize $pq$ using Algorithm \ref{alg:reorganizepq}
%\WHILE{\textit{pq} not empty}
\STATE $E^* = \emptyset$
\STATE $t = 0$
\REPEAT
\IF{$pq$ is empty}
\IF{$R$ is empty}
\STATE Break
\ENDIF
\STATE  $pq$ = Refill($L$)
\ENDIF
\STATE Extract edge $e$ with highest priority from $pq$ or by random sampling if $pq$ is empty or lower priority than default
\IF{Sufficient statistics of $e$ not already computed}
\STATE Compute sufficient statistics of $e$ \COMMENT{\footnotesize {cost: $O(N s_{\max}^2)$}}
\ENDIF
\STATE Compute score $s_e$. \COMMENT{\footnotesize {cost: $O(s_{\max}^2)$}}
\IF{$s_e > \lambda$ and $R$ not full}
\STATE Add $e$ to $R$ \COMMENT{\footnotesize {add $e$ if capacity not reached}}
\ELSIF{$s_e > \lambda$ and ($R$ is full and $R_{\min} < s_e$)}
\STATE Replace $R_{\min}$ by $e$. \COMMENT{\footnotesize {$R_{\min}$: minimum scroing edge in $R$}}
\STATE Place $R_{\min}$ in $L$.
\ELSE
\STATE Place $e$ in $L$
\ENDIF
\UNTIL{ $t=t_{\max}$}
\STATE Compute  $\tau$
\FOR{$e \in R \text{ s.t } s_e\geq \tau$} 
\IF{$e$ not adjacent to edges in $E^*$}  \label{nonadjacent}
\STATE $E^* \cup \{e\}$
\ENDIF
\ENDFOR
%\ENDWHILE
\end{algorithmic}
\caption{Activation Test}
\label{alg:Activation}
\end{algorithm*}

\begin{algorithm*}[h!]
\begin{algorithmic}[1]
\STATE Define EdgeNum as the maximum allowed number of edges to graft
\STATE Initialize $E = \emptyset$, $F= $ set of all possible edges
\STATE Initialize empty \textit{pq}  \label{initialize_pq}
\STATE AddedEges = 0
\STATE Fill $R$ to capacity
\WHILE{EdgeNum $>$ AddedEges}
\STATE Get set $E^*$ of edges to activate using Algorithm \ref{alg:Activation}
\IF{$E^*$ is empty}
\STATE Break
\ENDIF
\FOR{$e^*$ in $E^*$}
\STATE $E = E \cup \{e^*\}$; $F = F \backslash \{e^*\}$
\STATE AddedEges $ = $ AddedEges $ + 1$
\ENDFOR
\STATE Perform an optimization over active parameters.
\ENDWHILE
\end{algorithmic}
\caption{Best-Choice Edge Grafting}
\label{alg:OEG}
\end{algorithm*}

\section{Note on Initializing the Priority Queue}

Our experiments show that there is only a small practical benefit of using random sampling for default-priority edges. For the data sizes in our experiments, the $O(n^2)$ cost to instead initialize the full priority queue was negligible, since it is a one-time operation. The quadratic cost only becomes a practical bottleneck when it is repeated or compounded by the data size $N$, as is the case for edge grafting. Thus, our experiments report running times where our best-choice edge-grafting methods take an extra half second or less for the one-time initialization of the full priority queue. 

\section{Synthetic Data Generation Details}

We generate MRFs with different sizes and variables cardinality five. In particular, we generate structures of size 200, 400, and 600, with 498,500, 1,997,000, and 4,495,500 parameters, respectively.

We first construct random, scale-free structured graphs. To do so, we use a preferential-attachment model where we grow a graph by attaching new nodes, each with two edges that are preferentially attached to existing nodes with high node centrality. The algorithm produces $(2n-4)$ edges. For each node and each created edge, we sample their corresponding parameters from a normal distribution with a mean equal to $100$ and standard deviations $\sigma_v = 0.5$ (for nodes) and $\sigma_e = 1$ (for edges). This setting puts more emphasis on the edges (pairwise relationships between nodes) and helps avoid making the model overly dependent on only the unary potentials.

To generate data, we use a Gibbs sampler to generate a set of $20,000$ data points, which we randomly split as $19,000$ training data points and $1,000$ held-out testing data points. 

\paragraph*{Parameter tuning}
We use a grid search to detect the best combination of $\lambda$ and $\lambda_2$. We limit our search range to the set $\{10^{-4}, 10^{-3}, 10^{-2}, 10^{-1}, 1\}$ for $\lambda$ and the set $\{0.5, 0.75, 1, 1.5\}$ for $\lambda_2$. For each case, we choose the pair $(\lambda, \lambda_2)$ that produces the best recall and test NLPL for the baseline, i.e., edge grafting. It is worth noting that we did not notice a high sensitivity of the methods for different values of $ \lambda_2$, whereas smaller values of $\lambda$ produce spurious edges for different methods.

\end{document}